\documentclass[graybox]{svmult}
\usepackage[utf8]{inputenc}
\usepackage{type1cm}        
\usepackage{makeidx}         
\usepackage{graphicx}        
\usepackage{multicol}        
\usepackage[bottom]{footmisc}
\usepackage{booktabs}
\usepackage{multirow}
\usepackage{enumerate}

\usepackage{newtxtext}       %
\usepackage{newtxmath}       

\def\x{{\mathbf x}}														
\def\q{{\mathbf q}}														
\def\C{{\mathbf C}}														\def\E{\text{E}} 									
\def\W{{\mathbf W}}														
\def\ii{{\hat{\imath}}}												
\def\ij{{\hat{\jmath}}}												
\def\ik{{\hat{\kappa}}}												
\def\bH{\mathbb{H}}														
\def\bR{\mathbb{R}}														
\def\cL{{\cal L}}															
\newcommand{\tT}{^{\text{T}}} 								
\newcommand{\tH}{^{\text{H}}} 								
\DeclareMathOperator{\Tr}{Tr}								
\DeclareMathOperator{\var}{var}								
\begin{document}

\title*{Quaternion Generative Adversarial Networks} 
\author{Eleonora Grassucci, Edoardo Cicero and Danilo Comminiello}
\institute{Authors are with the Department of Information Engineering, Electronics and Telecommunications (DIET), Sapienza University of Rome, Via Eudossiana 18, 00184 Rome, Italy, \email{\{eleonora.grassucci, danilo.comminiello\}@uniroma1.it}. This work has been supported by ``Progetti di Ricerca Grandi'' of Sapienza University of Rome under grant number RG11916B88E1942F.
}

\maketitle

\abstract*{Latest Generative Adversarial Networks (GANs) are gathering outstanding results through a large-scale training, thus employing models composed of millions of parameters requiring extensive computational capabilities. Building such huge models undermines their replicability and increases the training instability. Moreover, multi-channel data, such as images or audio, are usually processed by real-valued convolutional networks that flatten and concatenate the input, often losing intra-channel spatial relations. To address these issues related to complexity and information loss, we propose a family of quaternion-valued generative adversarial networks (QGANs). QGANs exploit the properties of quaternion algebra, e.g., the Hamilton product, that allows to process channels as a single entity and capture internal latent relations, while reducing by a factor of 4 the overall number of parameters. We show how to design QGANs and to extend the proposed approach even to advanced models. We compare the proposed QGANs with real-valued counterparts on several image generation benchmarks. Results show that QGANs are able to obtain better FID scores than real-valued GANs and to generate visually pleasing images. Furthermore, QGANs save up to $75\%$ of the training parameters. We believe these results may pave the way to novel, more accessible, GANs capable of improving performance and saving computational resources.
}

\abstract{Latest Generative Adversarial Networks (GANs) are gathering outstanding results through a large-scale training, thus employing models composed of millions of parameters requiring extensive computational capabilities. Building such huge models undermines their replicability and increases the training instability. Moreover, multi-channel data, such as images or audio, are usually processed by real-valued convolutional networks that flatten and concatenate the input, often losing intra-channel spatial relations. To address these issues related to complexity and information loss, we propose a family of quaternion-valued generative adversarial networks (QGANs). QGANs exploit the properties of quaternion algebra, e.g., the Hamilton product, that allows to process channels as a single entity and capture internal latent relations, while reducing by a factor of 4 the overall number of parameters. We show how to design QGANs and to extend the proposed approach even to advanced models. We compare the proposed QGANs with real-valued counterparts on several image generation benchmarks. Results show that QGANs are able to obtain better FID scores than real-valued GANs and to generate visually pleasing images. Furthermore, QGANs save up to $75\%$ of the training parameters. We believe these results may pave the way to novel, more accessible, GANs capable of improving performance and saving computational resources.
}

\section{Introduction}
\label{sec:intro}
\noindent Generative models including generative adversarial networks (GANs) \cite{GoodfellowNIPS2014} and variational autoecoders (VAEs) \cite{KingmaARXIV2014} have been recently spectators of an increasing widespread development due to the massive availability of large datasets covering a large range of applications. The demand to learn such complex data distributions leads to define models far from the original approach of a simple GAN, which was characterized by fully connected layers and evaluated on benchmark datasets such as the MNIST \cite{GoodfellowNIPS2014}. Multiple pathways have been covered to improve GANs generation ability. A first branch aims at stabilizing the training process which is notoriously unstable, often leading to a lack of convergence. This includes constraining the discriminator network to be 1-Lipschitz by introducing a gradient penalty in the loss function, normalizing the spectral norm of the network weights or adding a consistency regularization \cite{Arjovsky2017WassersteinG, GulrajaniNIPS2017, Miyato2018SpectralNF, ZhangConsRegGAN2020}. 
Other significant improvements are gained by architectural innovations such as self-attention modules, flexible activation functions or style-based generator \cite{KarrasStyleGen2019, GrassucciFlexGAN2021, ZhangSAGAN2019, KarrasPG2018}. A crucial improvement in the quality of image generation has been brought by broadly scaling up the networks and involving wider batch sizes \cite{Brock2019LargeSG, KarrasSG22020, schonfeld2021unet}. Indeed, BigGAN closed the visual quality gap between GANs generated images and real-world samples in ImageNet \cite{Brock2019LargeSG}. Most of the latest GANs are somehow inspired to it. 

However, these impressive results come at the cost of huge models with hundred of millions of free parameters which require large computational resources. This drastically reduces the accessibility and the diffusion of these kind of models. Moreover, GANs are notorious fragile models, thus the training with this amount of parameters may result in unstable or less handy process.
Furthermore, when dealing with multidimensional inputs, such as images, 3D audio, multi-sensor signals or human-pose estimation, among others, real-valued networks break the original structure of the inputs. Channels are processed as independent entities and just concatenated in a tensor without exploiting any intra-channel correlation.

In order to address these limitations, neural networks in hypercomplex domains have been proposed. Among them, quaternion neural networks (QNNs) leverage the properties of the non-commutative quaternion algebra to define lower-complexity models and preserve relations among channels. Indeed, QNNs process channels together as a single entity, thus maintaining the original input design and correlation. Due to this feature, QNNs are able to capture internal relations while saving up to the $75\%$ of free parameters thanks to hypercomplex-valued operations, including the Hamilton product.

Encouraged by the promising results of other generative models in the quaternion domain \cite{GrassucciICASSP2021, Grassucci2021Entropy} and the need to make deep GANs more accessible, we introduce the family of quaternion generative adversarial networks (QGANs). QGANs are completely defined in the quaternion domain and, among other properties, they exploit the quaternion convolutions derived from the hypercomplex algebra \cite{ParcolletAIR2019, ParcolletICLR2019, GaudetIJCNN2018, ComminielloICASSP2019a} to improve the generation ability of the model while reducing the overall number of parameters. A first attempt to introduce quaternion convolutions in GANs has been recently made in \cite{Qgan2021Sfikas}. Here, we define the core-blocks of the quaternion generative adversarial framework that we use to formulate a vanilla QGAN. Then, we explain how to derive more advanced QGANs to prove the superior generation ability of the proposed approach in multiple image generation benchmarks. We show that the quaternion spectral normalized GAN (QSNGAN) is able to earn a better FID and a more pleasant visual quality of the generated images with respect to its real-valued counterpart thanks to the quaternion inner operations. Moreover, the proposed QSNGAN has just $25\%$ the number of free parameters with respect to the real-valued SNGAN.

We believe that these theoretical statements and empirical results lay the foundations for novel deep GANs in hypercomplex domains capable of grasping internal input relations while scaling down computational requirements, thus saving memory and being more accessible. To the best of our knowledge, this is the first time that a generative adversarial framework has been completely defined in a hypercomplex domain.

The contribution of this chapter is threefold: 
\begin{enumerate}[i)]
    \item we introduce the family of quaternion generative adversarial networks (QGANs) proving their enhanced generation ability and lower-complexity with respect to its real-valued counterpart on different benchmark datasets\footnote{The implementation of the QGANs is available online at \url{https://github.com/eleGAN23/QGAN}};
    \item we define the theoretically correct approach to apply the quaternion batch normalization (QBN) and redefine existing approaches as its approximations; 
    \item we propose and define the spectral normalization in the quaternion domain (QSN) proving its efficacy on two image generation benchmarks.
\end{enumerate}

The chapter is organized as follows. Section \ref{sec:qalg} presents the fundamental properties of quaternion algebra, while Section \ref{sec:qlearn} describes the quaternion adversarial framework and the quaternion-valued core blocks used in QGANs. Section \ref{sec:qgan} lays the foundations for the quaternion generative adversarial networks and presents a simple quaternion vanilla GAN and a more advanced and complex QGAN model. Section \ref{sec:exp} proves the effectiveness of the presented QGANs on a thorough empirical evaluation, and, finally, conclusions are drawn in Section \ref{sec:conc}.

\section{Quaternion Algebra}
\label{sec:qalg}
\noindent Quaternions are hypercomplex numbers of rank $4$, being a direct non-commutative extension of complex-valued numbers. The quaternion domain $\bH$ lies in a four-dimensional associative normed division algebra over real numbers, belonging to the class of Clifford algebras \cite{Ward1997}.
A quaternion is defined as the composition of one scalar element and three imaginary ones:

\begin{equation}
\label{eq:q}
    q = q_0 + q_1 \ii + q_2 \ij + q_3 \ik = q_0 + \mathbf{q}
\end{equation}

\noindent with $q_0, \; q_1, \; q_2, \; q_3 \in \bR$ and being $\ii = (1,0,0), \; \ij=(0,1,0), \; \ik=(0,0,1)$ unit axis vectors representing the orthonormal basis in $\bR^3$. A \textit{pure quaternion} is a quaternion without its scalar part $q_0$, resulting in the vector $\mathbf{q} = q_1 \ii + q_2 \ij + q_3 \ik$. As for complex numbers, also the quaternion algebra relies upon the relations among the imaginary components:

\begin{gather}
    \ii^2 = \ij^2 = \ik^2 = -1 \\
    \ii \ij = \ii \times \ij = \ik ; \; \ij \ik = \ij \times \ik = \ii ; \; \ik \ii = \ik \times \ii = \ij
\end{gather}

While the scalar product of two quaternions $q$ and $p$ is simply defined as the element-wise product $q \cdot p = q_0p_0 + q_1p_1 + q_2p_2 + q_3p_3$, quaternion vector multiplication, denoted with $ \times $, is not commutative, i.e., $\ii \ij \neq \ij \ii$. In fact:

\begin{equation*}
    \ii \ij = - \ij \ii ; \; \ij \ik = - \ik \ij ; \; \ik \ii = - \ii \ik.
\end{equation*}

Due to the non-commutative property, we need to introduce the quaternion product, commonly known as Hamilton product. We will see that Hamilton product plays a crucial role in neural networks. It is defined as:

\begin{equation}
	\begin{split}
		qp &= \left(q_0 + q_1\ii + q_2\ij + q_3\ik\right)\left(p_0 + p_1\ii + p_2\ij + p_3\ik\right) \\
		&= \left(q_0 p_0 - q_1 p_1 - q_2 p_2 - q_3 p_3\right) \\
		&+ \left(q_0 p_1 + q_1 p_0 + q_2 p_3 - q_3 p_2\right)\ii \\
		&+ \left(q_0 p_2 - q_1 p_3 + q_2 p_0 + q_3 p_1\right)\ij \\
		&+ \left(q_0 p_3 + q_1 p_2 - q_2 p_1 + q_3 p_0\right)\ik. 
	\end{split}
	\label{eq:qprod}
\end{equation}

The above product can be rewritten in a more concise form as:

\begin{equation}
\label{eq:qprod_concise}
    qp = q_0 p_0 - \mathbf{q} \cdot \mathbf{p} + q_0 \mathbf{p} + p_0 \mathbf{q} + \mathbf{q} \times \mathbf{p},
\end{equation}

\noindent where $q_0 p_0 - \mathbf{q} \cdot \mathbf{p}$ is the scalar element of the new quaternion in output and $q_0 \mathbf{p} + p_0 \mathbf{q} + \mathbf{q} \times \mathbf{p}$ is instead the vector part of the quaternion. From \eqref{eq:qprod_concise} it is easy to define a concise form of product for pure quaternions too:

\begin{equation}
    \mathbf{q}\mathbf{p} = - \mathbf{q} \cdot \mathbf{p} + \mathbf{q} \times \mathbf{p}.
\end{equation}

\noindent where the scalar product is the same as before for full quaternions and the vector product is $\mathbf{q} \times \mathbf{p} = (q_2 p_3 - q_3 p_2) \ii + (q_3 p_1 - q_1 p_3) \ij + (q_1 p_2 - q_2 p_1) \ik$. 

Similarly to complex numbers, the complex conjugate of a quaternion can be defined as:

\begin{equation}
    q^* = q_0 - q_1 \ii - q_2 \ij - q_3 \ik = q_0 - \mathbf{q}
\end{equation}

\noindent Also the norm is defined and it is equal to $|q| = \sqrt{q q^*} =  \sqrt{q_0^2 + q_1^2 + q_2^2 + q_3^2}$ that is the euclidean norm in $\bR^4$. Indeed, $q$ is said to be a \textit{unit quaternion} if $|q|=1$, as well as a \textit{pure unit quaternion} if $q^2 = -1$. Moreover, a quaternion $q$ is endowed with an inverse determined by:

\begin{equation*}
    q^{-1} = \frac{q^*}{|q|^2}.
\end{equation*}

\noindent Note that for unit quaternions, the relation $q^* = q^{-1}$ holds.

A quaternion has also a polar form: 

\begin{equation}
    q = \left|q\right|\left(\cos\left(\theta\right) + \mathbf{v}\sin\left(\theta\right)\right) = \left|q\right|e^{\mathbf{v}\theta}
\end{equation}
 
\noindent where $\theta \in \bR$ is the argument of the quaternion, $\cos\left(\theta\right) = q_0/\left\|q\right\|$,  $\sin\left(\theta\right) = \left\|\overline{q}\right\|/\left\|q\right\|$ and $\mathbf{v} = \overline{q}/\left\|\overline{q}\right\|$ is a pure unit quaternion.

Following, quaternions show interestingly properties when they can be interpreted as points and hyperplanes in $\mathbb{R}^4$. Among them, we find involutions, which are generally defined as self-inverse mappings or mappings that are their own inverse.
Quaternions have an infinite number of involutions \cite{ELL2007137} that can be generalized by the formula:

\begin{equation}
    q^{\mathbf{v}} = - \mathbf{v} q \mathbf{v}
\end{equation}

\noindent where $q$ is an arbitrary quaternion to be involved and $\mathbf{v}$ is any unit vector and the axis of the involution. Among the infinite involutions, the most relevant ones are the three perpendicular involutions defined as:

\begin{equation}
    \begin{split}
        &q^\ii = -\ii q \ii = q_0 + q_1 \ii - q_2 \ij - q_3 \ik \\
        &q^\ij = -\ij q \ij = q_0 - q_1 \ii + q_2 \ij - q_3 \ik \\
        &q^\ik = -\ik q \ik = q_0 - q_1 \ii - q_2 \ij + q_3 \ik
    \end{split}
\label{eq:inv}
\end{equation}

\noindent which are the first involutions identified \cite{Chernov1995} and they are crucial for the study of the second-order statistics of a quaternion signal, as we will see in the next section.

\section{Generative Learning in the Quaternion Domain}
\label{sec:qlearn}
In this section, we introduce the quaternion adversarial approach as well as the fundamental quaternion-valued operations employed to define the family of QGANs in next sections. It is worth noting that in a quaternion neural network each element is a quaternion, including inputs, weights, biases and outputs.

\subsection{The Quaternion Adversarial Framework}

\noindent Generative adversarial networks are built upon a minimax game between the generator network ($G$) and the discriminator one ($D$), as a special case of the concept initially proposed to implement artificial curiosity \cite{SchmidhuberMIT1991, SchmidhuberNEUNET2020}. They are trained in an adversarial fashion through the following objective function introduced in \cite{GoodfellowNIPS2014}:

\begin{equation}
\label{eq:gan_loss}
    \min_{G}\max_{D} V(D,G) = \E_{x\sim p_{\text{data}}(x)}\left\{\log D(x)\right\} \\
    + \E_{z\sim p_z(z)}\left\{\log (1-D(G(z)))\right\}
\end{equation}

\noindent where $p_{\text{data}}$ is the real data distribution and $p_z$ is the noise distribution. The two terms in the objective are two cross-entropies \cite{Gui2020ARO}. Indeed, the first term is the cross-entropy between $[1 \quad 0]\tT$ and $[D(x) \quad 1-D(x)]\tT$, whereas the second term is the cross-entropy between $[0 \quad 1]\tT$ and $[D(G(z)) \quad 1-D(G(z))]\tT$. 
In order to introduce the family of QGANs, first we need to delineate this adversarial approach in the hypercomplex domain. Thus, we define the cross-entropy function for quaternions which has to take the four components into account, as suggested in \cite{ParcolletAIR2019} for the quaternion mean squared error, by replacing real numbers with hypercomplex numbers and computing the operations element-wise. Thus, the quaternion cross-entropy (QCE) between the target quaternion $q$ and the estimated one $\tilde{q}$ can be defined as follows:

\begin{equation}
\label{eq:q_ce}
    \begin{split}
    \text{QCE}(q, \tilde{q}) = \frac{1}{N} \sum_{n=1}^{N} &\big[ q_0\log(\tilde{q}_0) + (1-q_0) \log(1-\tilde{q}_0) \\
    &+ q_1\log(\tilde{q}_1) + (1-q_1) \log(1-\tilde{q}_1) \\
    &+ q_2\log(\tilde{q}_2) + (1-q_2) \log(1-\tilde{q}_2) \\
    &+ q_3\log(\tilde{q}_3) + (1-q_3) \log(1-\tilde{q}_3) \big].
    \end{split}
\end{equation}

More in general, several objective functions proposed to train GANs can be redefined in the quaternion domain.
Among the most common ones, we find the Wasserstein distance with a gradient penalty that enforces the Lipschitz continuity of the discriminator, which is defined as follows \cite{Arjovsky2017WassersteinG, GulrajaniNIPS2017}: 

\begin{equation}
\label{eq:wgan}
    V(D,G) = \E_{x\sim p_{\text{data}}}\left\{D(x)\right\} - \E_{z\sim p(z)}\left\{D(G(z))\right\} \\
    - \lambda \E_{\hat{x}\sim p_{\hat{x}}} \left\{(||\nabla_{\hat{x}}D({\hat{x}})||_2-1)^2\right\}
\end{equation}

\noindent where the last term is the gradient penalty that is a regularization technique for the discriminator.

Other works \cite{Miyato2018SpectralNF, Chen2019SelfSupervisedGV} consider instead the hinge loss, which is given, respectively for the discriminator and the generator, by:

\begin{equation}
\label{eq:hinge_disc}
    V(D, \hat{G}) = \E_{x\sim p_{\text{data}}(x)}\left\{\min (0, \: - 1 + D(x))\right\} \\
    + \E_{z\sim p_z(z)}\left\{\min (0, \: - 1 - D(G(z)))\right\},
\end{equation}

\begin{equation}
\label{eq:hinge_gen}
    V(\hat{D},G) = - \E_{z\sim p_z(z)}\left\{\hat{D}(G(z)))\right\}.
\end{equation}

Being \eqref{eq:wgan} and \eqref{eq:hinge_disc} the composition of expected values and cross-entropies, both the definitions of the Wasserstein loss and of the hinge loss in the quaternion domain are straightforwardly derived by following the procedure shown for the adversarial loss in \eqref{eq:gan_loss}.

\subsection{Quaternion Fully Connected Layers}

\noindent In real-valued neural networks, fully connected layers are generally defined as:

\begin{equation}
\label{eq: mlp_layer}
    \mathbf{y}_{\text{r}} = \phi(\mathbf{W}_{\text{r}}\mathbf{x}_{\text{r}} + \mathbf{b}_{\text{r}})
\end{equation}

\noindent where $\mathbf{W}_{\text{r}}\mathbf{x}_{\text{r}}$ performs the multiplication between the weight matrix $\mathbf{W}_{\text{r}}$ and the input $\mathbf{x}_{\text{r}}$, $\mathbf{b}_{\text{r}}$ is the bias and $\phi(\cdot)$ is any activation function.
In order to define the same operation in the quaternion domain, we represent the  quaternion weight matrix as $\mathbf{W} = \mathbf{W}_0 + \mathbf{W}_1 \ii + \mathbf{W}_2 \ij + \mathbf{W}_3 \ik$, the quaternion input as $ \x = \x_0 + \x_1 \ii + \x_2 \ij + \x_3 \ik$ and the quaternion bias as $\mathbf{b} = \mathbf{b}_0 + \mathbf{b}_1 \ii + \mathbf{b}_2 \ij + \mathbf{b}_3 \ik$. Therefore, $\mathbf{W}\mathbf{x}$ in \eqref{eq: mlp_layer}, is performed by a vector multiplication between two quaternions, i.e., by the Hamilton product $\W \otimes \x$:

\begin{equation}
	\begin{split}
	\W \otimes \x &= \left(\W_{0}\x_{0} - \W_{1}\x_{1} - \W_{2}\x_{2} - \W_{3}\x_{3}\right)\\
	&+ \left(\W_{1}\x_{0} + \W_{0}\x_{1} - \W_{3}\x_{2} + \W_{2}\x_{3}\right)\ii \\
	&+ \left(\W_{2}\x_{0} + \W_{3}\x_{1} + \W_{0}\x_{2} - \W_{1}\x_{3}\right)\ij \\
	&+ \left(\W_{3}\x_{0} - \W_{2}\x_{1} + \W_{1}\x_{2} + \W_{0}\x_{3}\right)\ik. \\
    \end{split}
\label{eq:qmlp}
\end{equation}

Note that $\mathbf{W}$ has dimensionality $\frac{1}{4}|\mathbf{W}_{\text{r}}|$ since it is composed of four submatrices $\mathbf{W}_0, \mathbf{W}_1, \mathbf{W}_2$ and $\mathbf{W}_3$ each one with $1/16$ the dimension of $\mathbf{W}_{\text{r}}$. This is a key feature of QNNs since the results of the quaternion layer with product $\W \otimes \x$ has the same output dimension of the real-valued layer built upon $\W_{\text{r}} \mathbf{x}_{\text{r}}$ but with $1/4$ the number of parameters to train.
Note also that the submatrices are shared over each component of the quaternion input. The sharing allows the weights to capture internal relations among quaternion elements since each charcteristic in a component will have an influence in the other components through the common weights. In this way the relations among components are preserved and captured by the weights of the network which is able to process inputs without losing intra-channel information.
The bias $\mathbf{b}$ is then added with a sum component by component.
Finally, in QNNs the activation functions are applied to the input element-wise resulting in the so called \textit{split activation functions}. That is, suppose to consider a common Rectified Linear Unit (ReLU) activation function and the quaternion $\mathbf{z} = \W \otimes \x + \mathbf{b}$, the final result $\mathbf{y}$ of the layer will be:

\begin{equation}
    \mathbf{y} = \text{ReLU}(\mathbf{z}_0) + \text{ReLU}(\mathbf{z}_1) \ii + \text{ReLU}(\mathbf{z}_2) \ij + \text{ReLU}(\mathbf{z}_3) \ik.
\end{equation}

\subsection{Quaternion Convolutional Layers}
\label{sec:qconv}
\noindent Convolutional layers are generally applied to multichannel inputs, such as images. Supposing to deal with color images, real-valued neural networks break the structure of the input and concatenates the red, green and blue (RGB) channels in a tensor. Quaternion-valued convolutions, instead, preserve the correlations among the channels and encapsulates the image in a quaternion as \cite{ParcolletAIR2019, ParcolletICASSP2019a, QilinQCNN2019}:

\begin{equation}
    \x = 0 + R\ii + G\ij + B\ik
\label{eq:qimg}
\end{equation}

\begin{figure}
\centering
    \includegraphics[scale=0.6]{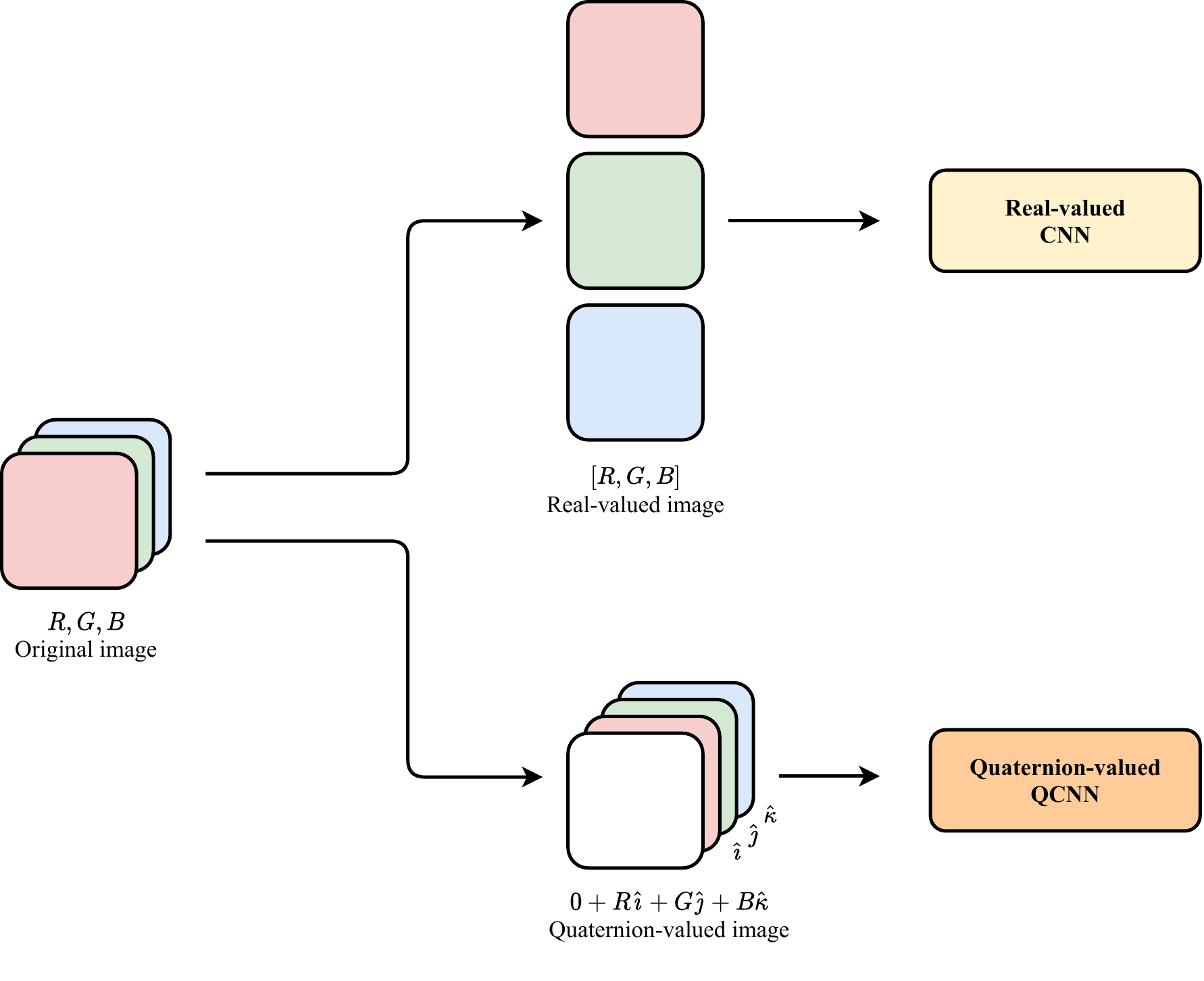}
\caption{Visual explanation of an $R,G,B$ image processed by real and quaternion-valued networks. On the left, the original three-channels image. The image can be processed in two ways: i) As a tensor of independent channels by a standard real-valued convolutional network as on the top of the figure. ii) As a single entity, encapsulating it in a quaternion, and considering internal relations among channels as quaternion-valued convolutional network does in the bottom of the figure. It is worth noting how the real-valued network does not consider any correlation among channels while quaternion ones preserve the relations among channels.}
    \label{fig:rgb_images}
\end{figure}

\noindent The image channels are the real coefficients of the imaginary units while the scalar part is set to $0$. Encapsulating channels in a quaternion allows to treat them as a single entity and thus to preserve intra-channels relations. A visual explanation of the quaternion representation of color images is depicted in Fig.~\ref{fig:rgb_images}.

Similarly to the definition of fully connected layers in the previous section, let us consider now a real-valued convolutional layer delineated by:

\begin{equation}
\label{eq:conv_layer}
    \mathbf{y} = \phi(\mathbf{W}_{\text{r}} * \mathbf{x}_{\text{r}} + \mathbf{b}_{\text{r}})
\end{equation}

\noindent where $*$ is the convolution operator.
Quaternion convolutional layers are built with the same procedure depicted for fully connected layers thus considering the Hamilton product instead of the standard vector multiplication. That is, the convolution operator $\mathbf{W}_{\text{r}} * \mathbf{x}_{\text{r}}$ is replaced for quaternion weights and inputs with

\begin{equation}
	\begin{split}
	\W * \x &= \left(\W_{0}*\x_{0} - \W_{1}*\x_{1} - \W_{2}*\x_{2} - \W_{3}*\x_{3}\right)\\
	&+ \left(\W_{1}*\x_{0} + \W_{0}*\x_{1} - \W_{3}*\x_{2} + \W_{2}*\x_{3}\right)\ii \\
	&+ \left(\W_{2}*\x_{0} + \W_{3}*\x_{1} + \W_{0}*\x_{2} - \W_{1}*\x_{3}\right)\ij \\
	&+ \left(\W_{3}*\x_{0} - \W_{2}*\x_{1} + \W_{1}*\x_{2} + \W_{0}*\x_{3}\right)\ik. \\
    \end{split}
\label{eq:qconv}
\end{equation}

\begin{figure}
    \centering
    \includegraphics[scale=0.65]{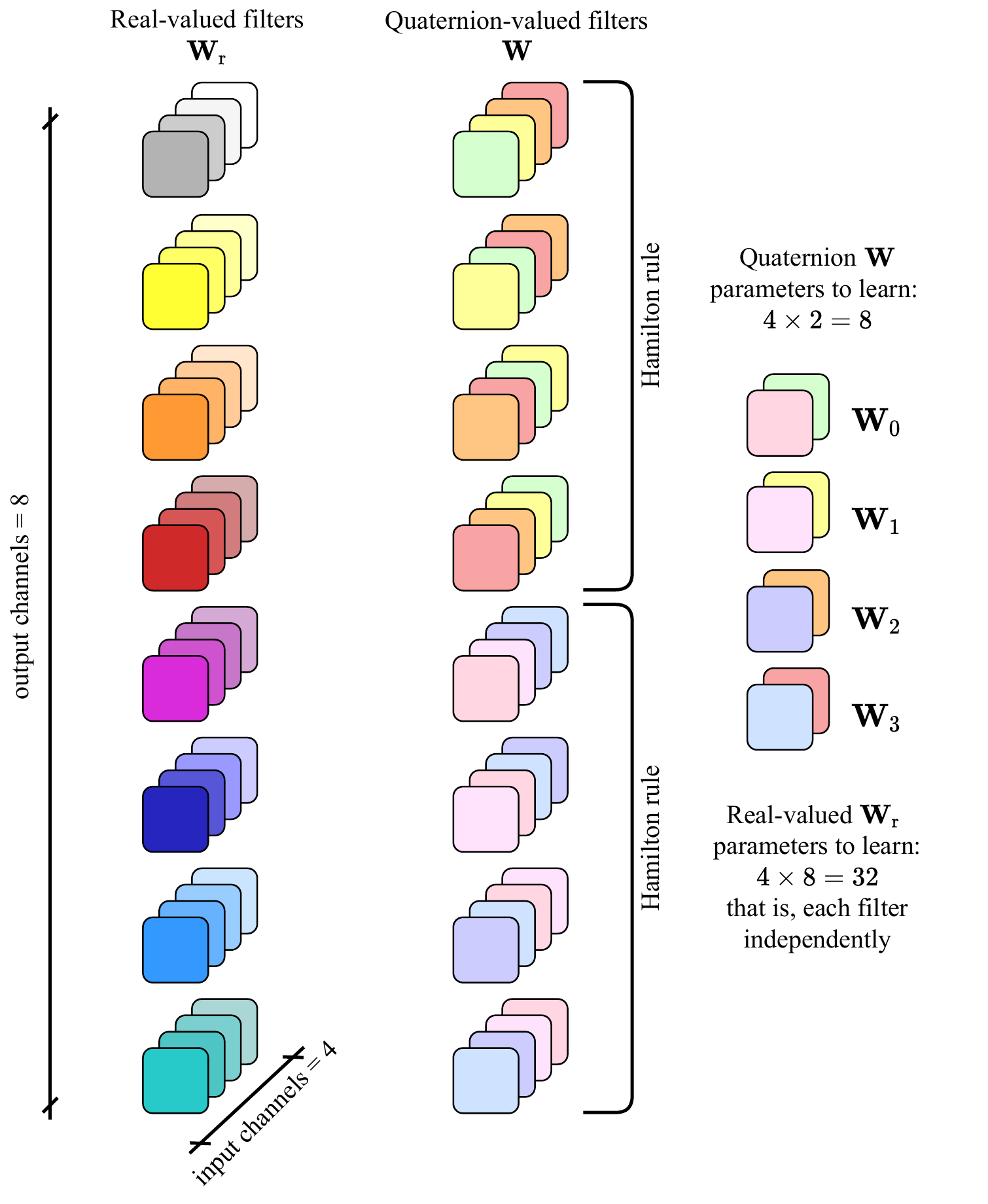}
    \caption{Real and quaternion convolution filters for a layer $l$ with $4$ channels in input and $8$ channels in output. On the left, the real-valued network has to learn each filter independently, thus resulting in $4 \times 8 = 32$ filters to learn. On the right, the quaternion filters are adjusted following the Hamilton product rule in \eqref{eq:qconv} and shared among the $4$ input components. As a result, the dimension of the weight matrix $\mathbf{W}$ is the same as $\mathbf{W}_{\text{r}}$, but the quaternion filters to learn are just $4 \times 2 = 8$, that is, the $25\%$ of the real-valued filters. The so-composed filters are then employed to perform convolution.}
    \label{fig:qconv}
\end{figure}

\noindent A visual explanation of the operation is shown in fig.\ref{fig:qconv}. While real-valued convolutional layer has to learn each filter independently, quaternion convolution allow the sharing of filters, thus reducing the number of free parameters to train.

Note that in convolutional networks the sharing weights are crucial to properly process channels. Indeed, the RGB channels of an image interact with each other by resulting in combined colors, such as yellow or violet, through a representation of pixels in the color space. Nonetheless, real-valued networks are not able to catch these interactions since they process input channels separately, while QCNNs not only preserves the input design but also capture these relations through the sharing of weights.
Actually, QCNNs perform a double learning: the convolution operator has the task of learning external relations among the pixels of the image, while the Hamilton product accomplishes the learning among the channels.
Furthermore, as for linear layers, QCNNs are built with $1/4$ the number of parameters with respect to their real-valued counterpart.

\subsection{Quaternion Pooling Layers}
\noindent Many neural networks make use of pooling layers, such as max pooling or average pooling, to extract high-level information and reduce input dimensions. As done before for previous layers in the quaternion domain, also this set of operations can be redefined in the quaternion domain.

The simplest examples of pooling in the hypercomplex domain are average and sum poolings. Indeed, applying these operations to each quaternion component, as done for split activation function, will not affect the final result \cite{QilinQCNN2019}. A different approach must be defined, instead, for max pooling. Indeed, the maximum of a single component is not guaranteed by the maximum of all the other components. In order to address this issue, a guidance matrix has to be introduced. As in \cite{QilinQCNN2019}, the matrix is built through the quaternion amplitude and keeps trace of the maximum position, which is then mapped back to the original quaternion matrix in order to proceed with the pooling computation. However, max pooling operations are rarely employed in GANs, thus we only make use of average and sum pooling in our experiments.

\subsection{Quaternion Batch Normalization}
\noindent Introduced in \cite{IoffeBN2015}, batch normalization (BN) has immediately became an ever-present module in neural networks. The idea behind BN is to normalize inputs to have zero mean and unit variance. This normalization helps the generalization ability of the network among different batches of training data and between train and test data distribution. 
Moreover, reducing the internal covariate shift remarkably improves the training speed, thus leading to a faster convergence of the model.
For these reasons, also QNNs are endowed with batch normalization. However, different versions of this method were proposed in literature. An elegant whitening procedure based on the standard covarince matrix is introduced in \cite{GaudetIJCNN2018}. In that paper, the Cholesky decomposition is used to compute the square root of the inverse of the covariance matrix, which is often intractable. The authors asserts that approach ensures zero mean, unit variance and decorrelation among components. However, the covariance matrix is not able to recover the complete second-order statistics in the quaternion domain \cite{TookSIGPRO2011} and the decomposition requires heavy matrix calculations and computational time \cite{HoffmanAlgebraNets2020}. Another remarkable approach is introduced in \cite{VecchiTIT2020}, where the input is standardized computing the average of the variance of each component. Nevertheless, describing the second-order statistics of a signal in the quaternion domain needs meticulous computations and the approach in \cite{VecchiTIT2020} is an approximation of the complete variance. Notwithstanding the approximation, this method allows to notably reduce computational time.

The proper theoretically procedure to reach a centered, decorrelated and unit-variance quaternion signal would be represented by performing a whitening procedure. Ideally, we should consider the covariance matrix and then decompose it to whiten the input in order to avoid computing the square root of the inverse which is often unfeasible. However, due to the interactions among components, second-order statistics for quaternion random variables are not completely described by the standard covariance matrix \cite{TookSIGPRO2011}. For this reason, the augmented covariance matrix should be considered instead. Such matrix is augmented with the complementary covariance matrices $\C_{\q\q^i}, \; \C_{\q\q^j}, \; \C_{\q\q^k}$ that are the covariance matrices of the quaternion with its three perpendicular involutions $\q^i, \q^j, \q^k$. Thus, the augmented covariance matrix, which completely characterizes the second-order information of the augmented quaternion vector $\tilde{\q}$, is defined as:

\begin{equation}
	\tilde{\C}_{\q\q} = \E\left\{\tilde{\q}\tilde{\q}\tH\right\} 
	= \left[{\begin{array}{*{20}c} {{\C}_{\q\q}} & {{\C}_{\q\q^\ii}} & {{\C}_{\q\q^\ij}} & {{\C}_{\q\q^\ik}}\\
	{{\C}_{\q\q^\ii}\tH} & {{\C}_{\q^\ii \q^\ii}} & {{\C}_{\q^\ii \q^\ij}} & {{\C}_{\q^\ii \q^\ik}}\\
	{{\C}_{\q\q^\ij}\tH} & {{\C}_{\q^\ij \q^\ii}} & {{\C}_{\q^\ij \q^\ij}} & {{\C}_{\q^\ij \q^\ik}}\\
	{{\C}_{\q\q^\ik}\tH} & {{\C}_{\q^\ik \q^\ii}} & {{\C}_{\q^\ik \q^\ij}} & {{\C}_{\q^\ik \q^\ik}}\\
	\end{array}} \right]
	\label{eq:augcovariance}
\end{equation}

\noindent where $(\cdot)\tH$ is the conjugate transpose operator. The formulation in \eqref{eq:augcovariance} recovers the complete statistical information of a general quaternion signal. Thus, the theoretically procedure should be delineated as:

\begin{equation}
    \overline{\mathbf{x}} = \tilde{\C}_{\q\q}^{-1/2}(\mathbf{x} - \E\left\{\mathbf{x}\right\})
\end{equation}

\noindent or substituting the inverse square root $\tilde{\C}_{\q\q}^{-1/2}$ with a decomposition of it. However, the construction of the augmented covariance matrix may be quite difficult and computational expensive due to the computation of each sub-covariance matrix. Moreover, $\tilde{\C}_{\q\q}^{-1/2}$ includes skew-symmetric sub-matrices \cite{TookSIGPRO2011}, which make the decomposition more difficult.

In order to simplify the calculation of \eqref{eq:augcovariance} and make it more feasible for practical applications, a particular case can be considered by leveraging the $\mathbb{Q}$-properness property \cite{ViaTIT2010, TookSIGPRO2011, GrassucciICASSP2021}. The $\mathbb{Q}$-properness entails that the quaternion signal is not correlated with its involutions, implying vanishing complementary covariance matrices, i.e., $\C_{\q\q^i} = \C_{\q\q^j} = \C_{\q\q^k} = 0$. Also, for $\mathbb{Q}$-proper random variables the following relation holds:

\begin{equation}
    \var\left\{\q_c\right\} = \E\left\{\q_c^2 \right\} = \sigma^2, \; \ \ \ \ c=\{0, 1, 2, 3\}
\label{eq:qproper}
\end{equation}

\noindent Thus, considering a $\mathbb{Q}$-proper quaternion, the covariance in \eqref{eq:augcovariance} becomes:

\begin{equation}
	\tilde{\C}_{\q\q} = \E\left\{\tilde{\q}\tilde{\q}\tH\right\} 
	= \left[{\begin{array}{*{20}c} {{\C}_{\q\q}} & {\mathbf{0}} & {\mathbf{0}} & {\mathbf{0}}\\
	{\mathbf{0}} & {{\C}_{\q^\ii \q^\ii}} & {\mathbf{0}} & {\mathbf{0}}\\
	{\mathbf{0}} & {\mathbf{0}} & {{\C}_{\q^\ij \q^\ij}} & {\mathbf{0}}\\
	{\mathbf{0}} & {\mathbf{0}} & {\mathbf{0}} & {{\C}_{\q^\ik \q^\ik}}\\
	\end{array}} \right]
	= 4\sigma^2 \mathbf{I}
	\label{eq:augcovariance_proper}
\end{equation}

Assuming $\mathbb{Q}$-properness for a random variable  saves a lot of calculations and computational costs. Notwithstanding the theoretical correctness of the above defined approach, quaternion batch normalization (QBN) techniques adopted so far in the literature relies in some approximations.

We assume the input signal is $\mathbb{Q}$-proper, thus we consider the covariance in \eqref{eq:augcovariance_proper} and build the normalization as follows:

\begin{equation}
    \overline{\mathbf{x}} = \frac{\mathbf{x} - \mu_q}{\sqrt{\var\left\{\mathbf{x}\right\} + \epsilon}} = \frac{\mathbf{x} - \mu_q}{\sqrt{4\sigma^2 + \epsilon}}
\label{eq:q_normal}
\end{equation}

\noindent where $\mu_q$ is the quaternion input mean value, which is a quaternion itself, and it is defined as:

\begin{equation}
    \mu_q = \frac{1}{N} \sum_{n=1}^N q_{0,n} + q_{1,n} \ii + q_{2,n} \ij + q_{3,n} \ik = \bar{q}_0 + \bar{q}_1 \ii + \bar{q}_2 \ij + \bar{q}_3 \ik.    
\end{equation}




The final output is computed as follows:

\begin{equation}
    \text{QBN}(\mathbf{x}) = \gamma \overline{\mathbf{x}} + \beta
\label{eq:qbn}
\end{equation}

\noindent where $\beta$ is a shifting quaternion parameter and $\gamma$ is a scalar parameter.






In conclusion, the QBN proposed by \cite{GaudetIJCNN2018} is an elegant approximation, nevertheless it is not able to catch the complete second-order statistics information, while requiring heavy computations \cite{HoffmanAlgebraNets2020}. Thus, we believe that considering $\mathbb{Q}$-proper signals, which are indeed very frequent, is a good approximation which also extremely reduces the computational requirements. For our experiments, we adopt the method represented by \eqref{eq:qbn}.

\subsection{Quaternion Spectral Normalization}
\label{sec:qsn}

\noindent Among the wide variety of proposed techniques to stabilize GANs traning, the spectral normalization (SN) \cite{Miyato2018SpectralNF} is one of the most widespread method. Previously, the crucial importance of having a Lipschitz-bounded discriminator function was introduced in \cite{Arjovsky2017WassersteinG, GulrajaniNIPS2017}. Lately, it was proved that no restriction on the discriminator space leads to the \textit{gradient uninformativeness} problem \cite{ZhouLipGan2019}. This means that the gradient of the optimal discriminative function has no information about the real distribution, thus providing useless feedbacks to the generator. Forcing a function to be Lipschitz continuous means controlling how fast it increases and bound the gradients, thus mitigating gradient explosions \cite{ZhouLipGan2019, GoukLipNN2021}. In \cite{Arjovsky2017WassersteinG}, a method based on weight clipping was proposed to force the discriminator to be 1-Lipschitz. Later, such approach has been improved by adding a gradient penalty (GP) that constraints the gradient norm to be at most 1 \cite{GulrajaniNIPS2017}. The latter method is reproposed in several state-of-the-art GANs and combined with other regularization techniques to improve performance, as suggest \cite{Kurach2019ALS}. However, being built on the gradients with respect to the inputs, the gradient penalty cannot impose a regularization outside the support of the fake and real data distribution. Moreover, it requires consistent computations. The spectral normalization, instead, directly operates on the weights of the network being free of the support limit and its computations is faster than other methods \cite{Miyato2018SpectralNF}. It aims at controlling the Lipschitz constant of the discriminator by constraining the spectral norm of each layer.

A generic function $f$ is $K$-Lipschitz continuous if, for any two points $x_1$, $x_2$, the following property holds:

\begin{equation}
    \frac{\left\| f\left(x_1\right) - f\left(x_2\right)\right\|}{\left| x_1 - x_2 \right|} \le K
\end{equation}

\noindent being $\left\| \cdot \right\|$ the $l_2$ norm.
The Lipschitz norm $\left\|f\right\|_{\text{Lip}}$ of a function $f$ is equal to $\sup_x \sigma(\nabla f(x))$, where $\sigma(\cdot)$ is the spectral norm of the matrix in input, that is, the largest singular value of the matrix.

For a generic linear layer $f(h) = \mathbf{W}\mathbf{x} +\mathbf{b}$, the Lipschitz norm is:

\begin{equation}
    \left\|f\right\|_{\text{Lip}} = \sup_h \sigma (\nabla f(h)) = \sup_h \sigma(\mathbf{W}) = \sigma(\mathbf{W})
\label{eq:sigma_w}
\end{equation}

\noindent Assuming the Lipschitz norm of each layer activation being equal to 1, constraint that is satisfied for many popular activation functions including ReLU and Leaky ReLU \cite{Miyato2018SpectralNF}, we can apply the Lipschitz bound to the whole network by following $\left\|f_1 \circ f_2\right\|_{\text{Lip}} \le \left\|f_1\right\|_{\text{Lip}} \cdot \left\|f_2\right\|_{\text{Lip}}$.

Finally, the SN is defined as

\begin{equation}
    \bar{W}_{SN}(\mathbf{W}) = \frac{\mathbf{W}}{\sigma(\mathbf{W})}
\label{eq:sn}
\end{equation}

\noindent and it ensures that the weight matrix $\mathbf{W}$ always satisfies the constraint $\sigma(\mathbf{W})=1$. In \cite{Miyato2018SpectralNF} the authors underline that applying the original singular value decomposition algorithm to compute $\sigma(\W)$ may result in an extremely heavy algorithm. To address the computational complexity, they suggest to estimate the largest singular value via the power iteration method.

In order to control the Lipschitz constraint in a QGAN, in this section we explore two methods to define the spectral normalization in the quaternion domain. 
A first approach aims at normalizing the weights $\W$ by operating on each submatrix $\W_0, \; \W_1, \; \W_2, \; \W_3$ independently, by computing the spectral norm separately. That is, through the power iteration method as above, we compute $\sigma_0(\W_0), \; \sigma_1(\W_1), \; \sigma_2(\W_2), \; \sigma_3(\W_3)$ and then normalize each submatrix with the corresponding norm. This method forces each submatrix to have spectral norm equal to 1. However, it never takes the whole weight matrix $\W$ into account. Moreover, the relations among the components of the quaternion matrix is not considered, losing the characteristic property of QNNs. 

The second method, similarly to the real-valued SN, normalizes the whole matrix $\W$ together, by imposing the constraint to the complete matrix and not to the singular submatrices. Therefore, the spectral norm is computed by taking the complete weight matrix into account and considering the relations among the quaternion components. However, while the spectral norm is computed as in \eqref{eq:sigma_w}, the normalization step is applied differently from the SN in \eqref{eq:sn}. Instead of normalizing the whole matrix as in \eqref{eq:sn}, being the weight matrix $\W$ designed by a composition of the submatrices $\W_0, \; \W_1, \; \W_2, \; \W_3$, we can leverage this quaternion setup to save computational costs and normalize each submatrix $\W_0, \; \W_1, \; \W_2, \; \W_3$. The normalized submatrices $\bar{\W}_{0,QSN}, \; \bar{\W}_{1,QSN}, \; \bar{\W}_{2,QSN}, \; \bar{\W}_{3,QSN}$ will result in a normalized weight matrix $\bar{W}_{QSN}(\mathbf{W})$ with a more efficient computation than normalizing the full matrix $\W$.

An empirical comparison between the two methods is reported in Section \ref{sec:exp}. We investigate the two techniques in a plain QGAN and prove that the latter approach is stabler and gains better performance in both the datasets considered. We deem it more appropriate both theoretically and empirically and we use it in our further experiments. From now on, we refer to such approach as the quaternion spectral normalization (QSN).





\subsection{Quaternion Weight Initialization}
\label{sec:qwi}
\noindent Weight initialization has often a crucial role in networks convergence and in the reduction of the risk of vanishing or exploding gradients \cite{GlorotInit2010}. This procedure becomes even more important when dealing with quaternion weights. Indeed, due to the interactions among the elements of the quaternion, the initialization cannot be random nor component-aware. For these reasons, an appropriate initialization has to be introduced.

First, consider a weight matrix $\W$ with $\E\left\{|\W|\right\} = 0$. The initialization is based on a normalized pure quaternion $u$ generated for each weight submatrix from a uniform distribution in $[0, 1]$. By using the polar form of a quaternion, we can define the initialization of the weight matrix as

\begin{equation}
    \W = |\W|e^{u \theta} = |\W|(\cos(\theta) + u \sin(\theta))
\end{equation}

\noindent where each matrix component is initialized as

\begin{equation}
    \begin{split}
        \W_0 &= \phi \cos(\theta) \\
        \W_1 &= \phi u_1 \sin(\theta) \\
        \W_2 &= \phi u_2 \sin(\theta) \\
        \W_3 &= \phi u_3 \sin(\theta)
    \end{split}
\end{equation}

\noindent where the angle $\theta$ is randomly generated in the interval $[-\pi, \pi]$ and $\phi$ is randomly sampled in the interval of the standard deviation around zero $[-\sigma, \sigma]$. The standard deviation is set according to the initialization method chosen, either \cite{GlorotInit2010} or \cite{HeInit2015}. In the first case, we set $\sigma = 1/\sqrt{2(n_{in} + n_{out})}$ whereas in the latter we set $\sigma = 1/\sqrt{2 n_{in}}$. In both the equations, $n_{in}$ is the number of neurons in the input layer and $n_{out}$ the number of neurons in the output layer. The variance of $\W$ can be written as:

\begin{equation}
    \var\left\{\W\right\} = \E\left\{|\W|^2\right\} - \E\left\{|\W|\right\}^2.
\label{eq:qvar}
\end{equation}

However, similarly to the QBN in the previous section, in order to reduce the computations, the component $\E\left\{|\W|\right\}^2$ can be considered equal to $0$ \cite{ParcolletICLR2019, ParcolletAIR2019}. This is equivalent to considering a $\mathbb{Q}$-proper quaternion signal whose augmented covariance matrix has off-diagonal elements equal to $0$ and trace equal to $4\sigma^2$. Consequently, the variance is computed by considering only the first term of \eqref{eq:qvar} as:

\begin{equation}
    \var\left\{\W\right\} = \E\left\{|\W|^2\right\} = 4\sigma^2
\end{equation}

\subsection{Training}

The forward phase of a QNN is the same as its real-valued counterpart. Therefore, the input flows from the first to the last layer of the network. It may be interesting to note that in eq.~\eqref{eq:qmlp} the order of the weight and the input can be inverted, thus changing the output of the product, resulting in an inverted QNN \cite{ParcolletAIR2019, ParcolletICLR2019}.
For what concerns the backward phase, it worth mentioning that the gradient of a general quaternion loss function $\cL$ is computed for each component of the quaternion weight matrix $\W$ as in the ensuing equation:

\begin{equation}
    \frac{\delta \cL}{\delta \W} = \frac{\delta \cL}{\delta \W_0} + \frac{\delta \cL}{\delta \W_1} \ii + \frac{\delta \cL}{\delta \W_2} \ij + \frac{\delta \cL}{\delta \W_3} \ik.
\end{equation}

Then, the gradient is propagated back following the chain rule. Indeed, as defined in \cite{ParcolletAIR2019}, the backpropagation of quaternion neural networks is just an extension of the method for their real-valued counterpart. Consequently, QNNs can be easily trained as real-valued networks via backpropagation.

\section{GAN Architectures in the Quaternion Domain}
\label{sec:qgan}
\noindent The previous section described the main blocks and the framework to build and train a GAN in the quaternion domain. In this section we go further, presenting the complete definition of a plain QGAN in Subsection \ref{sec:vanilla_qgan} and of an advanced state-of-the-art QGAN composed of complex blocks in Subsection \ref{sec:adv_qgan}.
First, in order to setting up a QGAN, each input, weight, bias and output has to be manipulated to become a quaternion. Therefore, weight matrices are initialized as composed by the four submatrices, similarly to \eqref{eq:qmlp} and \eqref{eq:qconv}. Real-valued operations such as multiplications or convolutions in the networks are replaced with their quaternion counterparts, completing the redefinition of the layers in the quaternion domain. The input is handled as a quaternion and processed as a single entity. For images, a pure quaternion is considered as in \eqref{eq:qimg}, while for other kind of multidimensional signals, the scalar part is considered too. The initialization of the weights is then applied following the description in Section \ref{sec:qwi}. This accurate definition of QGANs grants to design a model with a fewer number of free parameters with respect to the same real-valued model and consequently to save memory and computational requirements.

\subsection{Vanilla QGAN}
\label{sec:vanilla_qgan}
\noindent In the original GAN \cite{GoodfellowNIPS2014}, both the generator ($G$) and the discriminator ($D$) are defined by fully connected layers. Due to the limited expressivity of this design with complex data such as images, in \cite{Radford2016UnsupervisedRL} the authors propose to replace dense layers with more suitable operations for this kind of data and to build $G$ and $D$ by stacking several convolutional layers. State-of-the-art GANs are based on the deep convolutional GAN (DCGAN) \cite{Radford2016UnsupervisedRL}. In particular, the DCGAN increases the spatial dimensionality by means of transposed convolutions in the generator and decreases it in the discriminator with convolutions. Furthermore, this architecture defines batch normalization in every layer except for the last layer of $G$ and for the first layer of $D$, in order to let the networks learn the correct statistics of the data distribution. 

By redefining the DCGAN in the quaternion domain (QDCGAN) it is possible to explore the potential of the quaternion algebra in a simple GAN framework. The QDCGAN generator is defined by an initial quaternion fully connected layer and then by interleaving quaternion transposed convolutions with quaternion batch normalization and split ReLU activation functions except for the last layer which ends up with a split Tanh function. The discriminator has the same structure of the generator but with quaternion transposed convolutions replaced by quaternion convolutions to decrease the dimensionality and with a final fully connected quaternion layer that returns as output the real/fake decision by means of a sigmoid split activation. The QDCGAN, as its real-valued counterpart, optimizes the original loss in \eqref{eq:gan_loss}.

\begin{figure}
    \centering
    \includegraphics[scale=0.7]{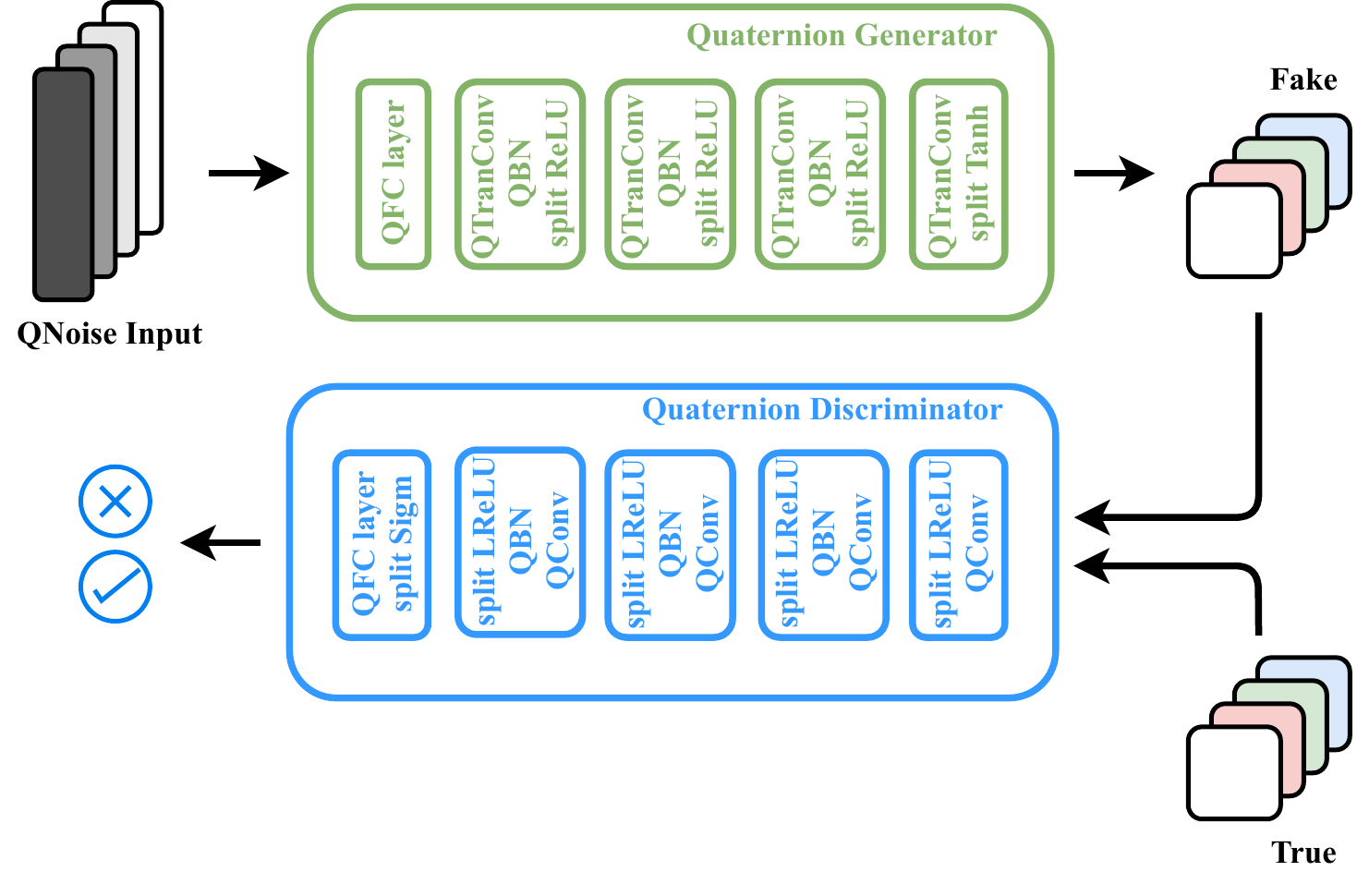}
    \caption{Quaternion Vanilla GAN architecture. Each parameter including inputs, weights and outputs is a quaternion. The generator (green network) takes a quaternion noise signal and generates a batch of quaternion images with four channels. The discriminator tries to distinguish between fake and real quaternion samples exploiting the properties of quaternion algebra.}
    \label{fig:qdcgan}
\end{figure}

\subsection{Advanced QGAN}
\label{sec:adv_qgan}

The above presented Vanilla QGAN is just a plain example to give a general idea on how to build GANs in the quaternion domain. In this section, we consider a more advanced model, the spectral normalized GAN (SNGAN) \cite{Miyato2018SpectralNF} and we present the steps to define its quaternion counterpart.

\begin{figure}[t]
    \centering
    \includegraphics[scale=0.6]{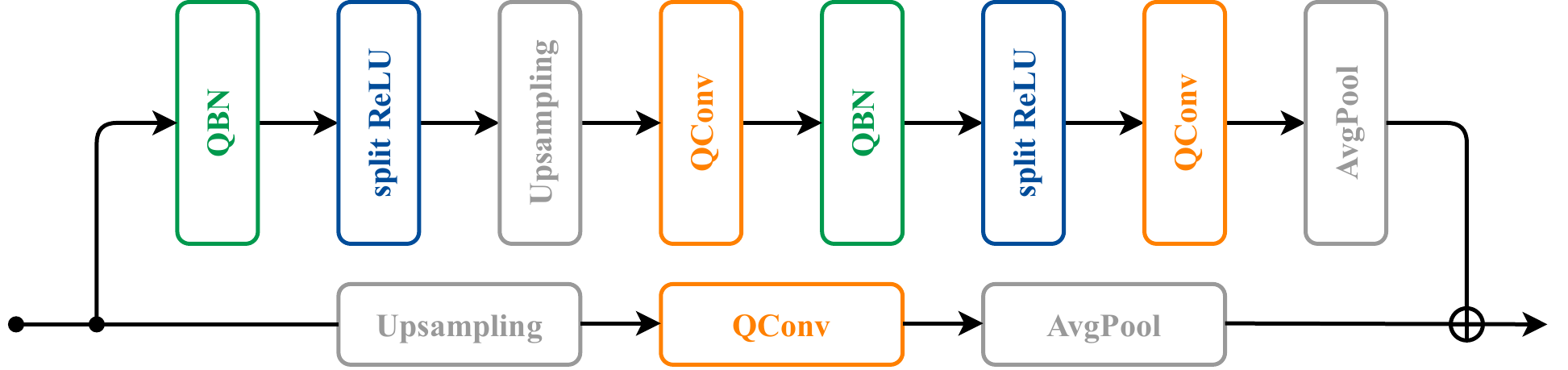}
    \caption{Quaternion residual block (QResBlock) architecture inspired by \cite{Miyato2018SpectralNF} and redefined in the quaternion domain. QBN is omitted in the discriminator network and replaced by QSN. Grey blocks means they are used exclusively in teh generator or in the discriminator. The generator considers the umpsampling steps in the residual and in the shortcut pass while the discriminator the average pooling ones, except for the last residual block of the discriminator which keep the dimension invariant.}
    \label{fig:resblock}
\end{figure}

\begin{figure}[t]
    \centering
    \includegraphics[scale=0.6]{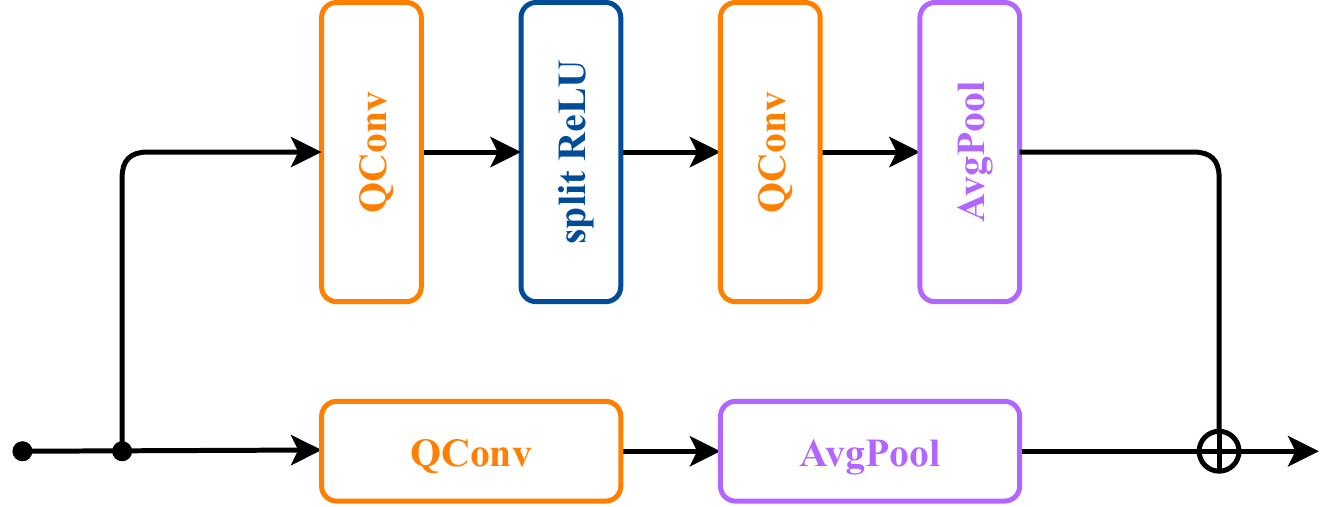}
    \caption{First discriminator quaternion residual block (First QResBlock) with quaternion convolutions and average pooling layers to downsample the input.}
    \label{fig:1resblock}
\end{figure}

The quaternion spectral normalized GAN (QSNGAN) is trained in an adversarial fashion through the hinge loss defined in \eqref{eq:hinge_disc} and \eqref{eq:hinge_gen} for the discriminator and generator respectively, as suggested in \cite{Miyato2018SpectralNF, Chen2019SelfSupervisedGV}. The overall architecture of the model is inspired by \cite{Chen2019SelfSupervisedGV}. Both the generator and discriminator networks are characterized by quaternion convolutional layers in order to leverage the properties of the Hamilton product. To mitigate the vanishing gradient problem and obtain better performance, a series of residual blocks with upsampling in the generator and downsampling in the disciminator can be adopted \cite{Miyato2018SpectralNF}. A scheme of the residual block of the proposed QSNGAN is depicted in Fig.~\ref{fig:resblock}. The discriminative network plays a crucial role in GANs training, thus it is more complex with respect to the generator network. It takes in input the two sets of quaternion images with four channels in a first residual block, as illustrated in Fig.~\ref{fig:1resblock}. The output of the block is the decision on whether they come from the fake or real distribution. 


In order to guarantee a fair comparison with the SNGAN, we consider a real-valued noise signal in input to the generator and handle it with an initial real-valued fully connected layer. The output of the first layer is then encapsulated in a quaternion signal with a procedure similar to the one considered in Subection~\ref{sec:qconv} to handle colored images. The signal is then processed by the quaternion generator up to the last layer, which generates the four-channel fake image. The original SNGAN considers batch normalization in the generator and spectral normalization in the discriminator. We keep the same structure and consider the proposed QBN in \eqref{eq:qbn} for the first network and the QSN introduced in Section \ref{sec:qsn} for the discriminator. In particular, we exploit the QSN with spectral norm computed over the whole weight matrix, which is theoretically better and ensures stabler results. 

\begin{figure}[t]
    \centering
    \includegraphics[scale=0.65]{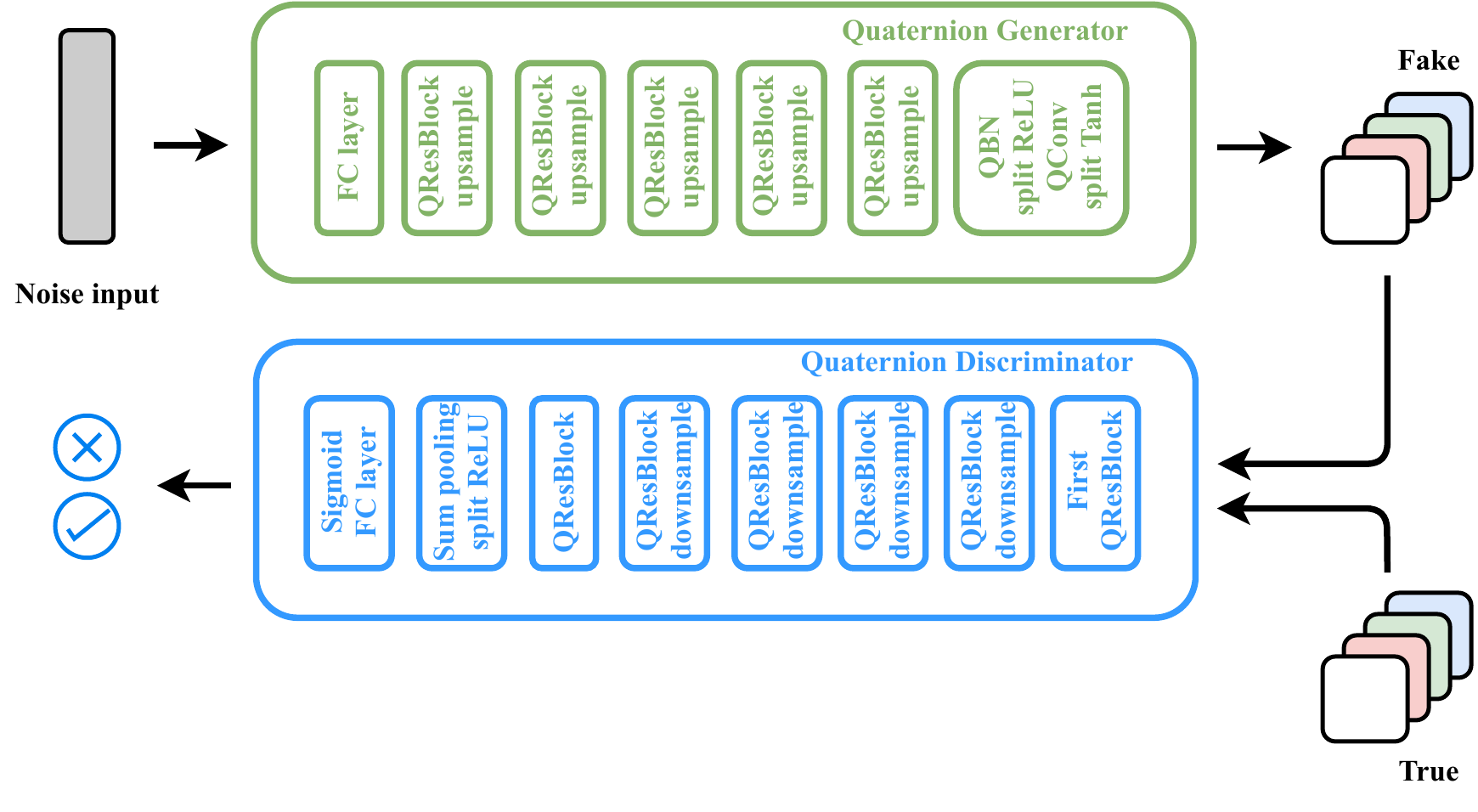}
    \caption{QSNGAN architecture schema. The generator network (top) takes in input a real-valued signal, processes it with a a fully connected layer and then encapsulates it in a quaternion signal. The residual blocks are depicted in Fig.~\ref{fig:resblock}. The generator outputs a quaternion-valued sample of images that, together with a sample from the real distribution, goes to the input of the discriminator network (bottom). It handles the samples through a series of residual blocks (the first one is illustrutade in Fig.~\ref{fig:1resblock}, the other ones in Fig.~\ref{fig:resblock}) up to the last layer which outputs the real or fake decision.}
    \label{fig:qsngan}
\end{figure}

The definition of the SNGAN in the quaternion domain allows to save parameters, as we will explore in the next section. Moreover, the QSNGAN, processing the channels as a single entity through the quaternion convolutions based on the Hamilton product, is able to capture the relations among them and to capture any intra-channel information, which the SNGAN, conversely, loses. The latter property turns into an improved generation ability by the QSNGAN that properly grasps the real data distribution. The architecture of the proposed QSNGAN is reported in Fig.~\ref{fig:qsngan}. In the scheme, the forward phase flows from left to right for the top network (quaternion generator) and from right to left for the second network (quaternion discriminator).

\subsection{Evaluation metrics}
In order to evaluate the performance of the generative networks, we consider two objective metrics, the Fréchet Inception Distance (FID) \cite{HeuselFID2017} as main metric, as it is more consistent with human evaluations, and the Inception Score (IS) \cite{SalimansIS2016}. The Fréchet inception distance embeds the generated and the real samples into the Inception convolutional features and models the two distributions as Gaussian signals evaluating the empirical means $\mu_g, \mu_{\text{data}}$ and covariances $\C_g, \C_{\text{data}}$ and then computes the Fréchet distance as:

\begin{equation}
    \text{FID}(p_g, p_{\text{data}}) = ||\mu_g - \mu_{\text{data}} || + \Tr(\C_g + \C_{\text{data}} -2(\C_g\C_{\text{data}})^{1/2}
\end{equation}

\noindent where $\Tr\left(\cdot\right)$ refers to the trace operation. Being the FID a distance between real and fake distributions, the lower the FID value, the better the generated samples.

Instead, the IS considers the inception model to get the conditional distribution $p(y|x)$ of the generated samples. IS expects the conditional distribution to have low entropy since the images represent meaningful objects, while the marginal distribution $p(y)$ should have high entropy due to the diversity among the samples. It is defined as:

\begin{equation}
    \text{IS}(p_g) = \exp\left(\E_{x \sim p_g} \left\{\text{KL}[p(y|x)||p(y)]\right\}\right)
\end{equation}

\noindent where $\text{KL}$ is the Kullback-Leibler divergence. Conversely to the FID, higher IS values stands for better generated samples. However, IS has some drawbacks since it does not consider the true data distribution and, moreover, it is not able to detect mode collapse, thus we consider the FID score as main metric and the IS in support to it. 

\section{Experimental Evaluation}
\label{sec:exp}
\noindent In order to evaluate the effectiveness of our proposed approach, we conduct a collection of experiments on the unsupervised image generation task. We take two datasets into account: the CelebA-HQ \cite{KarrasPG2018}, which contains 27k images for training and 3k images for testing, and the 102 Oxford Flowers, which contains approximately 7k images for training and a few less then 1k images for testing. We reshape the samples of both the dataset to $128 \times 128$ and then test the real-valued SNGAN and the proposed QSNGAN. We use the Adam optimizer and keep the same hyper-parameters fixed as in \cite{Miyato2018SpectralNF}, i.e., learning rate equal to $0.0002$, and the optimizer parameters equal to $\beta_1=0.0$, $\beta_2=0.9$. We just vary the number of critic iterations, considering two experiments with critic iterations equal to 1 and then equal to 5 in order to better investigate the behavior of our QSNGAN, which may have a different balance between generator and discriminator networks with respect to the SNGAN. In every experiment, we fix the batch size to 64 and we perform 100k training iterations for the CelebA-HQ and 50k for the 102 Oxford Flowers. We have also considered to endow the SNGAN and the QSNGAN with a gradient penalty, as in \eqref{eq:wgan}, but we did not notice any improvement in the experiments, thus meaning that both the SN and the QSN adequately control the discriminator to be Lipschitz continuous.

The QSNGAN generator is a quaternion convolutional network as in Fig.~\ref{fig:qsngan}. The initial fully connected layer, which takes the noise of size 128 in input, is composed of $4 \times 4 \times 1024$ neurons. The following quaternion residual blocks illustrated in Fig.~\ref{fig:resblock} stack 1024, 512, 256, 128 and 64 filters. This means that, as an example, the first residual block is built by interleaving QBNs, split ReLUs and quaternion convolutions with 1024 kernels and an upsampling module with scale factor equal to 2. Further, at the end of the last residual connection, we stack a QBN, a split ReLU activation function and a final quaternion convolutional layer of dimension 64 to refine the output image, which is then passed to a split Tanh function to bound it in the interval $[-1, 1]$. For each quaternion convolution, we fix the kernel size to 3 and the stride and the padding to 1. Conversely, the shortcut in the residual block is composed of an upsampling module and a quaternion convolution with kernel size equal to 1 and null padding. The network built through this procedure has less than 10M of free parameters with respect to the 32M parameters of its real-valued counterpart. This means that the checkpoint for inference saves more than the $70\%$ of disk memory, as shown in Table \ref{tab:params}.

\begin{table}[t]
\centering
\caption{Summary of number of networks parameters and memory requirements for real-valued SNGAN and its quaternion-valued counterpart QSNGAN models for CelebA-HQ. The proposed method saves more than the $70\%$ of total free parameters and memory disk for model checkpoints.}
\begin{tabular}{p{2cm}p{2.4cm}p{2.4cm}p{2.4cm}p{1.8cm}}
\hline\noalign{\smallskip}
 Model    & \#Params G & \#Params D & \#Params Tot & Disk Memory$^*$ \\
\noalign{\smallskip}\svhline\noalign{\smallskip}
 SNGAN  & 32,150,787    & 29,022,213    & 61,173,000      &  $\sim$115 GB \\
 QSNGAN & 9,631,204     & 7,264,901     & 16,896,105      &  $\sim$35 GB  \\
\noalign{\smallskip}\hline\noalign{\smallskip}

\end{tabular}\\
\label{tab:params}
$^*$ Generator checkpoint for inference.
\end{table}

The QSNGAN discriminator is still a quaternion convolutional network as in Fig.~\ref{fig:qsngan}, but it is slightly more complex. At the beginning, the real images are encapsulated in a quaternion as depicted in Subsection~\ref{sec:qconv}, resulting in a batch of four-channel images. Obviously, the images generated by the generator network are already comprised of four channels and defined as quaternions.

The first residual block of the discriminator in Fig.~\ref{fig:1resblock} is a spectrally-normalized quaternion convolution block with 64 $3 \times 3$ filters and split ReLU activation functions. The shortcut, instead, as for the generator network, is a $1 \times 1$ quaternion convolution with padding equal to 0. In this case, however, both the residual and the shortcut part ends with a $2 \times 2$ split average pooling. The images flow then to a stack of five residual blocks built as in Fig.~\ref{fig:resblock} with, respectively, 128, 256, 512, 1024 and 1024 filters. Nevertheless, the residual section of each block has a split average pooling to operate downsampling and the shortcut is comprised of a quaternion convolution and another average pooling. The downsampling procedure is applied in each residual block except for the last one, which is a refiner and leaves the dimensionality unchanged. Every weight is normalized through the QSN introduced in Subsection~\ref{sec:qsn}. The configurations for kernel size, stride and padding are the same of the generator. At the end of the residual block stack, we apply a split ReLU and a global sum pooling before passing the batch to the final spectrally-normalized fully connected layer which, by means of a sigmoid, returns the real/fake decision. As for the generator, also the quaternion discriminator allows to save parameters while learning the internal relations among channels. This saving is underlined in Table~\ref{tab:params}, which reports the exact number of parameters for the quaternion model and the real-valued one. The quaternion GAN can obtain equal or better results when trained with less parameters since it leverages the properties of quaternion algebra, including the Hamilton product, that allow to capture also the relations among channels and catch more information on the input. Consequently, the training procedure needs less parameters to learn the real distribution and to generate images from it.

The objective evaluation is reported in Table~\ref{tab:results}. We perform the computations of FID and IS on the test images (3k for the CelebA-HQ and slightly less than 1k for the 102 Oxford Flowers). As shown in Table~\ref{tab:results}, the proposed method stands out in the generation of samples from both the dataset according to the metrics considered. Moreover, the two QSNGANs with critic iterations 1 and 5 score a lower FID with respect to the best configuration of the SNGAN model. The proposed method performs better with one critic per generator iterations, while the real-valued model fails with this configuration. Overall, the QSNGAN seems to be more robust to the choice of the critic iterations with respect to the SNGAN, which is more fragile. The IS strengthen the results obtained with the FID, as it reports higher scores for the proposed method in every dataset.

\begin{table}[t]
\caption{Results summary for the $128 \times 128$ CelebA-HQ and 102 Oxford Flowers datasets. The proposed QSNGAN obtains a lower FID in each dataset considered. The vlaues of the IS support the FID results. According to the objective metrics, the proposed QSNGAN generates more visually pleasant and diverse samples with respect to the real-valued baseline counterpart. The QSNGAN seems to be more robust to the choice of the hyper-parameter regarding the discriminator iterations (Critic iter) while the real-valued model fails when changing the original setting which fixes the parameter equal to 5.}
\centering
\begin{tabular}{@{}lcccccccc@{}}
\hline\noalign{\smallskip}
       &                & \multicolumn{2}{c}{FID $\downarrow$}                                 & \multicolumn{2}{c}{IS $\uparrow$}                                                                                           & \multicolumn{2}{c}{}    \\ \midrule
Model  & Critic iter    & CelebA-HQ        & 102 Oxford Flowers                   & CelebA-HQ                                          & 102 Oxford Flowers                                          &                      &  \\
SNGAN  & 1              & $>$ 200$^*$     &       $>$ 200$^*$                     &     $<$ 2.000 $^*$                              &     2.797 $\pm$ 0.196         & \multicolumn{1}{c}{} &  \\
       & 5              & 34.483          & \multicolumn{1}{c}{\underline{165.058}}          & \multicolumn{1}{c}{\underline{2.032 $\pm$ 0.062}} & \multicolumn{1}{c}{\underline{2.977 $\pm$ 0.146}}          &                      &  \\
\noalign{\smallskip}\hline\noalign{\smallskip}
QSNGAN & 1              & \textbf{29.417} &  175.484                                    & \textbf{2.249 $\pm$ 0.164}            &       2.754 $\pm$ 0.256                                                      & \multicolumn{1}{c}{} &  \\
       & 5              & \underline{33.068}    & \multicolumn{1}{c}{\textbf{115.838}} & \multicolumn{1}{c}{}      2.026 $\pm$ 0.082                         & \multicolumn{1}{c}{\textbf{3.000 $\pm$ 0.141}} &                      & \\
\noalign{\smallskip}\svhline\noalign{\smallskip}
\end{tabular}
\\
\label{tab:results}
$^*$ Discriminator collapses and training fails, thus metrics results are not comparable.
\end{table}

\begin{figure}[t]
    \centering
    \includegraphics[scale=0.402]{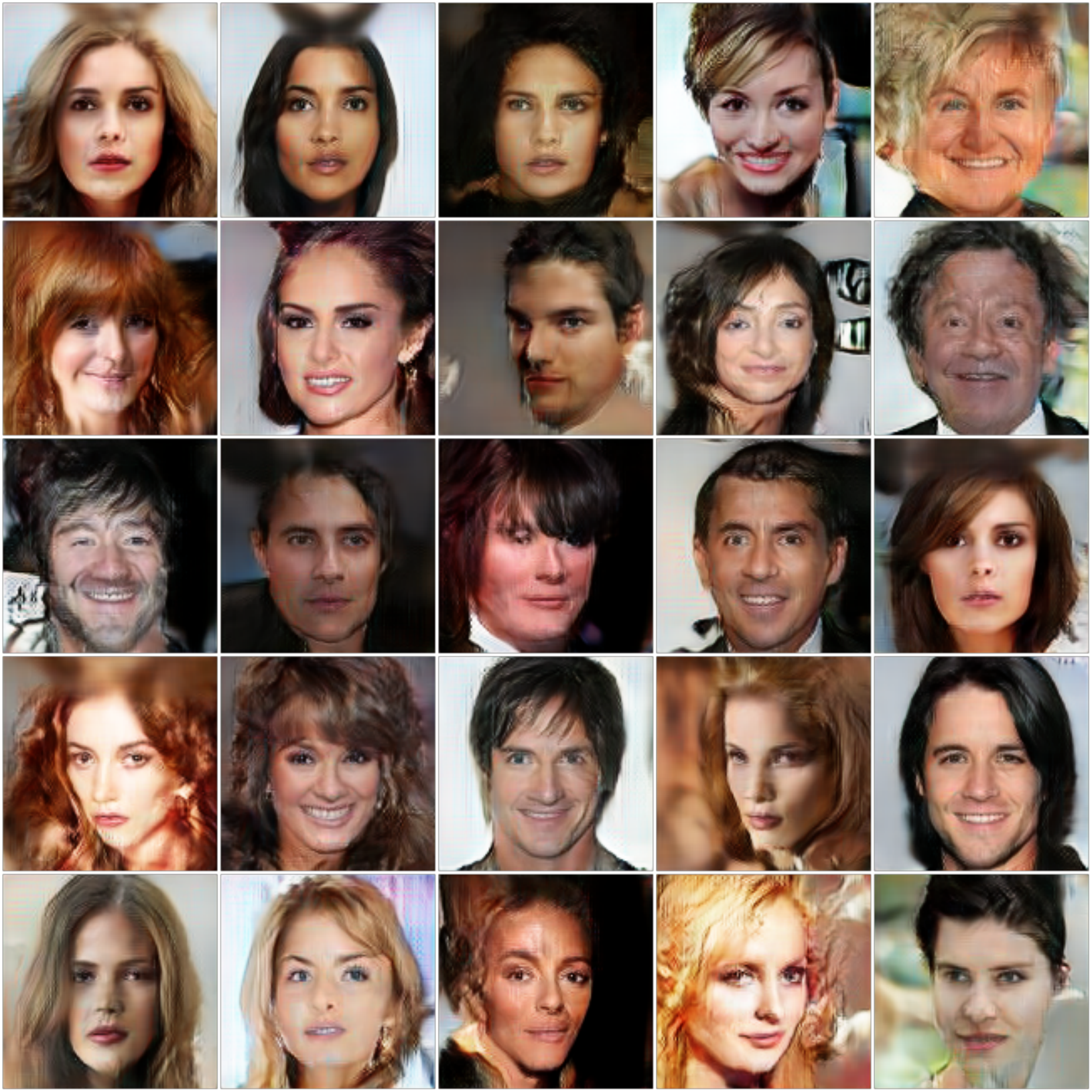}
    \caption{Randomly generated samples from the real-valued SNGAN on the CelebA-HQ dataset after 100k training iterations. Sometimes this model fails to detect border attributes such as hair and neck which may fade on the background. Indeed, only few samples seem to be visually pleasant while in some other cases the network fails to generate likable images.}
    \label{fig:res_celeba}
\end{figure}

\begin{figure}[t!]
    \centering
    \includegraphics[scale=0.402]{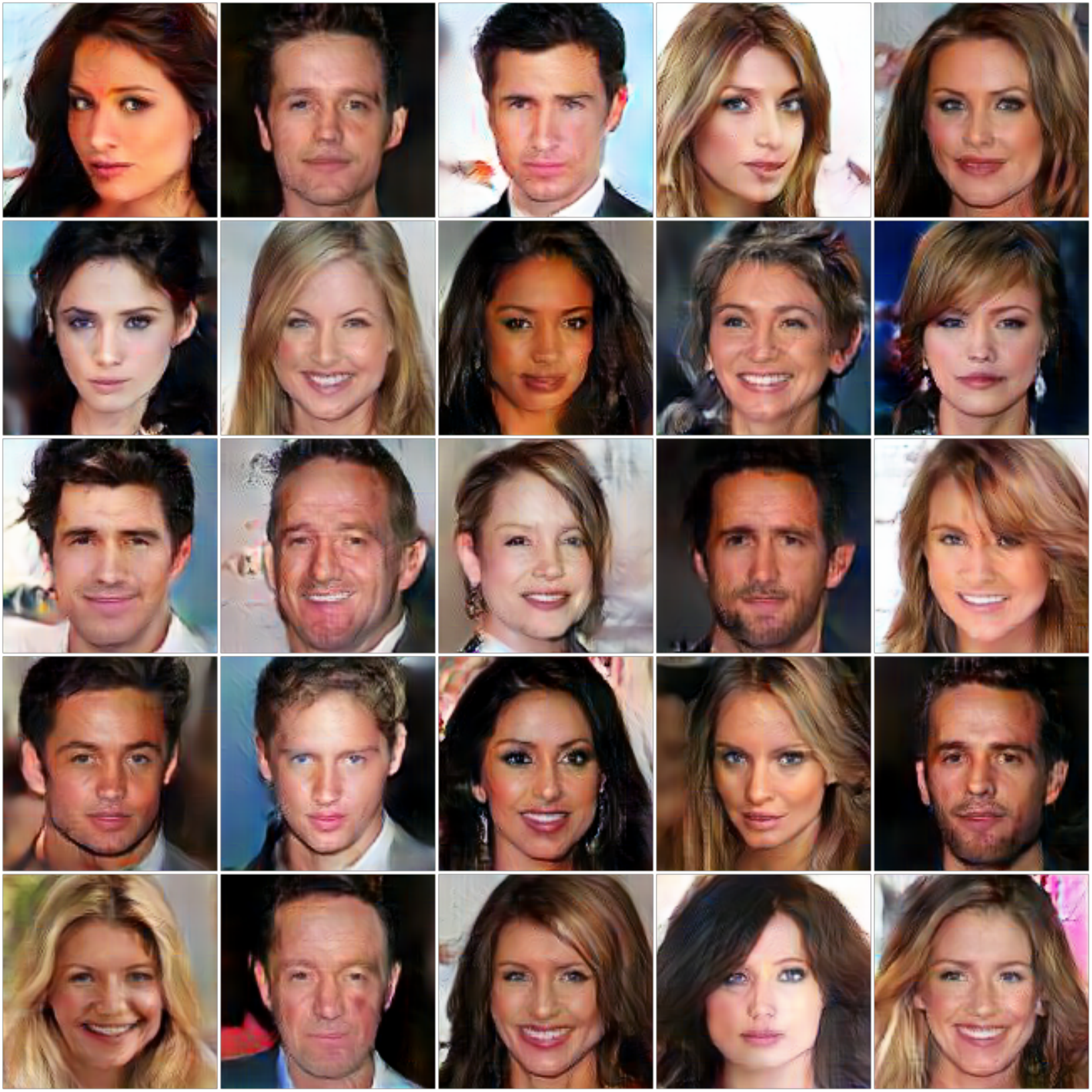}
    \caption{Randomly generated samples from our QSNGAN on the CelebA-HQ dataset after 100k training iterations. These images are part of the test samples which gained a FID of 29.417 and IS 2.249 $\pm$ 0.164. The proposed method is able to generate visually pleasant images, well distinguishing the background from the face. Moreover, we do not observe mode collapse as samples have different attributes such as genre, hair color, pose and smile, among others.}
    \label{fig:Qres_celeba}
\end{figure}

\begin{figure}[t]
    \centering
    \includegraphics[scale=0.4]{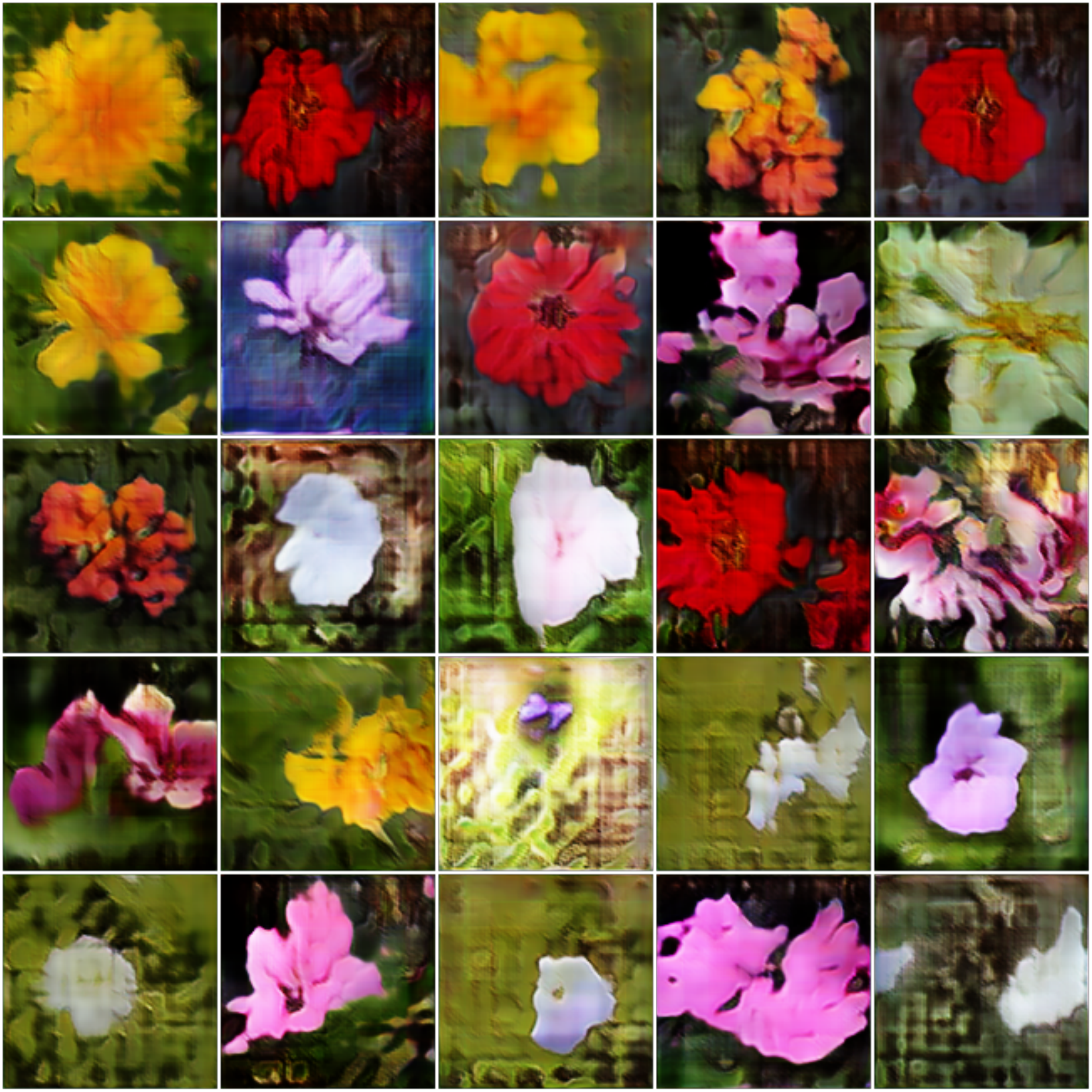}
    \caption{Randomly generated samples from the SNGAN model on the 102 Oxford Flowers dataset after 50k training iterations. SNGAN misleads some pixels in the images and depicted objects are not always distinguishable.}
    \label{fig:res_flowers}
\end{figure}

\begin{figure}[t]
    \centering
    \includegraphics[scale=0.4]{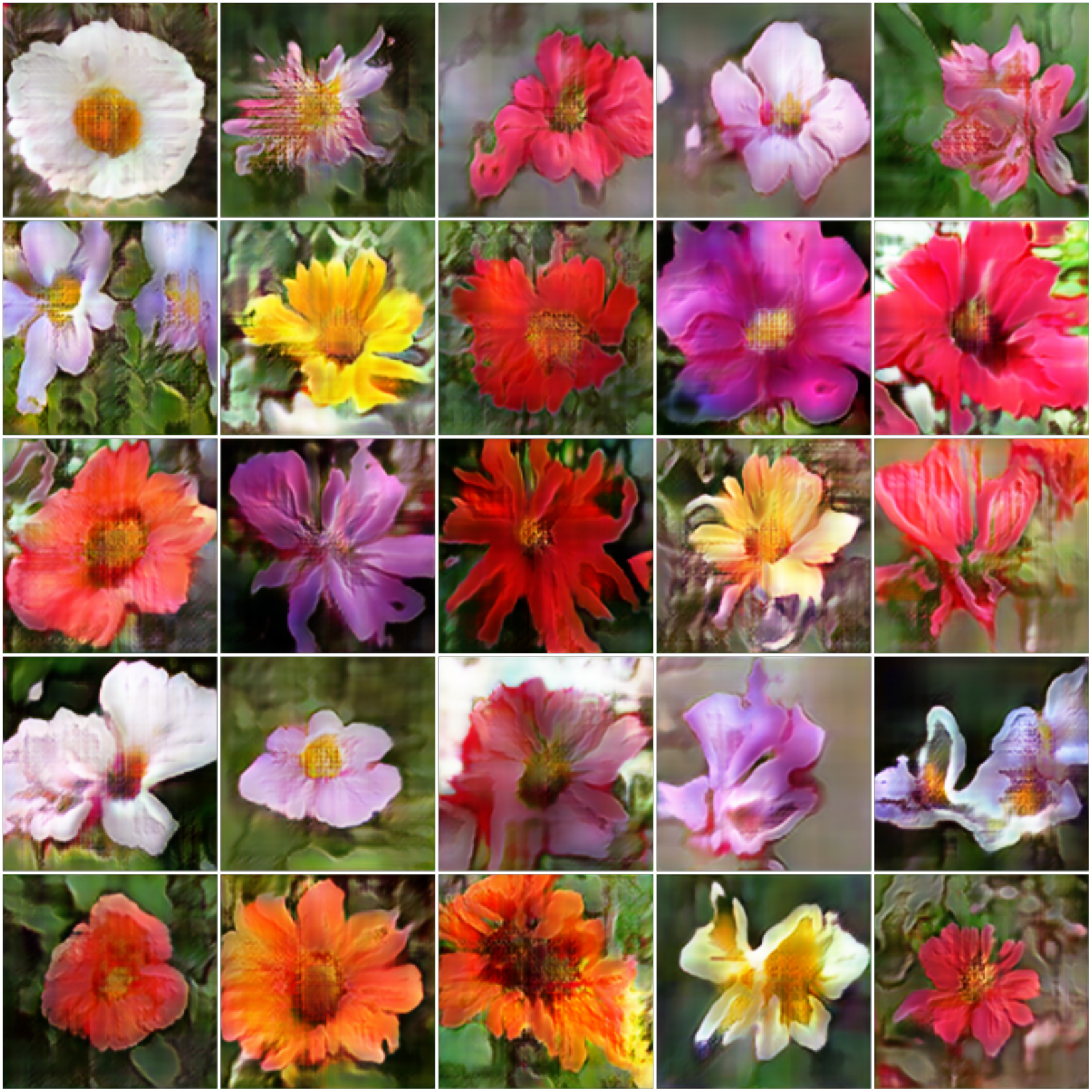}
    \caption{Randomly generated samples from the proposed QSNGAN on the 102 Oxford Flowers dataset after 50k training iterations. Flowers contain many different colors shades and most of the objects are clearly defined. This set of figures sow the improved generation ability of our proposed method with respect to its real-valued counterpart.}
    \label{fig:Qres_flowers}
\end{figure}

The visual inspection of the generated samples underlines the improved ability of our QSNGAN. Figure~\ref{fig:res_celeba} and Figure~\ref{fig:Qres_celeba} show a randomly selected $128 \times 128$ batch of generated images for the real-valued SNGAN and the proposed QSNGAN, respectively. On one hand, SNGAN seems to be quite unstable and prone to the input noise, thus alternating some good quality images with bad generated ones. Overall, the SNGAN is not always able to distinguish the background from some parts of the character, sometimes confusing attributes such as the neck or the hair as part of the environment, and letting them vanishing. On the other hand, the QSNGAN sample in Fig.~\ref{fig:Qres_celeba} shows visually pleasant images, with a clear distinction between subject and background. It also shows a higher definition of faces attributes, including the most difficult ones, such as eyebrows, beard or skin shades. In addition, colors seem to be more vivid and samples are diverse in terms of pose, genre, expression, and hair color, among others. 
Concerning the second dataset, the generated samples for the SNGAN are shown in Fig.~\ref{fig:res_flowers}, while the batch from the QSNGAN is reported in Fig.~\ref{fig:Qres_flowers}. As it is clear from Table~\ref{tab:results}, the results for this dataset are preliminary but encouraging. Even in this case the proposed approach gains a lower FID and a higher IS than the real-valued model. Additionally, in SNGAN samples pixels are evident and often misleading, thus confusing the flower object with the colored background. On the other hand, the images generated from our QSNGAN contain more distinct subjects. Furthermore, the proposed method better catches every color shade thanks to the quaternion algebra properties, which allow the network learning internal relations among channels without losing intra-channel information.

In conclusion, the proposed quaternion-valued QSNGAN shows an improved ability in capturing the real data distribution by leveraging the quaternion algebra properties in each experiment we conduct. It can generate better and more vivid samples according to visual inspections and to objective metrics with respect to its real-valued counterpart. Furthermore, the proposed method has less than the $30\%$ of free parameters with respect to the SNGAN which also has worse generation performance.

\subsection{Evaluation of Spectral Normalization Methods}
\noindent This section reports the tests we conduct to evaluate the two quaternion spectral normalization methods described in Subsection~\ref{sec:qsn}. In order to investigate the performance of the normalizing approaches, we validate two smaller models with respect to the ones introduced in the previous subsection on the CIFAR10 and STL10 datasets. CIFAR10 contains 50k $32 \times 32$ images for training and 10k for testing while STL10 has 105k $48 \times 48$ images in the train split and 8k in the test one.

We examine three different configurations: the first one does not involve any QSN method, thus the discriminator network is not constrained to be 1-Lipscithz. We run this experiment in order to check the effectiveness of the spectral normalization methods that we propose. The second configuration applies a split computation of the spectral norm for each quaternion component and normalize each weight submatrix $\W_0, \; \W_1, \; \W_2, \; \W_3$ independently. The last approach computes the spectral norm of the whole weight matrix and uses it to normalize each component. Respectively, we refer to these methods as No QSN, QSN Split and QSN Full.

To assess the performance, we build the same SNGANs presented in \cite{Miyato2018SpectralNF} by redefining them in the quaternion domain. We adopt the quaternion core residual blocks we define in the previous section and in Fig.~\ref{fig:resblock}, while reducing the model dimension. For CIFAR10, we set up a generator with the initial linear layer $4 \times 4 \times 256$ and then pile up three quaternion residual blocks, each one with 256 filters. As before, we end up with a stack of QBN, split ReLU and a quaternion convolution with a final split Tanh to generate the $32 \times 32$ images in the range $[-1, 1]$. The discriminator, in which the QSN methods act in each layer, begins with a first residual block (Fig.~\ref{fig:1resblock}) with 128 filters and then proceeds with three blocks composed of 128 kernels. As in Fig.~\ref{fig:qsngan}, the network ends with a global sum pooling and a fully connected layer with sigmoid to output the decision probability. The so-defined QSNGAN for CIFAR10 is comprised of less than 2M parameters. It is worth noting that the real-valued counterpart presented in \cite{Miyato2018SpectralNF} has more than 5M of free parameters.

The model to generate the $48 \times 48$ STL10 images is deeper than the previous one and is composed of 5,545,188 parameters. The structure is the same but it contains an initial layer of $6 \times 6 \times 512$ and then the residual blocks with 256, 128 and 64 filters. The final refiner quaternion convolutional layers has 64 kernels. The discriminator, instead, has one residual blocks more than the model for CIFAR10 and the filters are, respectively from the first to the last block, 64, 128, 256, 512, 1024 with a final 512 fully connected layer with sigmoid.

As we can see in Table~\ref{tab:qsn_tab}, the unbounded model with no QSN fails in generating images from both CIFAR10 and STL10. Indeed, the FID is much higher than the other approaches. This proves the effectiveness of the proposed QSN full method which computes the spectral norm of each layer taking all the components into account. As a matter of fact, the proposed approach is capable to generate improved quality images in every experiment we conduct.

\begin{table}[t]
\caption{Summary results for comparison of the two quaternion spectral normalization methods depicted in Section \ref{sec:qlearn}. We consider the SNGAN proposed in \cite{Miyato2018SpectralNF} as baseline to define two simple models in the quaternion domain and then test the different QSN approaches. QSN Split refers to the first method that normalizes the submatrices independently while QSN Full stands for the normalization of the whole weight matrix together. No QSN is a model without any spectral normalization method. While the latter fails, the QSN Full generates better images according to the FID in both datasets.}
\centering
\begin{tabular}{@{}lcccc@{}}
\hline\noalign{\smallskip}
          & \multicolumn{2}{c}{FID $\downarrow$}           & \multicolumn{2}{c}{IS $\uparrow$}                              \\ \midrule
Config    & CIFAR10         & STL10           & CIFAR10                   & STL10                   \\
No QSN    & 70.312          &  91.567         & 4.031 $\pm$ 1.327 &  \underline{4.744 $\pm$ 0.643}          \\
QSN Split & \underline{35.417}          & \underline{75.112}          & \textbf{4.7128 $\pm$ 1.270} & 4.455 $\pm$ 0.092           \\
QSN Full  & \textbf{31.966} & \textbf{59.611} & \underline{4.317 $\pm$ 0.951}            & \textbf{4.987 $\pm$ 0.485} \\ 
\noalign{\smallskip}\hline\noalign{\smallskip}
\end{tabular}
\label{tab:qsn_tab}
\end{table}

\section{Conclusions}
\label{sec:conc}
\noindent In this paper we introduce the family of quaternion-valued GANs (QGANs) that leverages the properties of quaternion algebra. We have rigorously defined each core block employed to build the proposed QGANs, including the quaternion adversarial framework. Moreover, we have provided a meticulous experimental evaluation on different image generation benchmarks to prove the effectiveness of our method. We have shown that the proposed QGAN has an improved generation ability with respect to the real-valued counterpart, according to the FID and IS metrics and to a visual inspection. Moreover, our method saves up to the $75\%$ of free parameters. We believe that these results lay the foundations for novel deep GANs, thus capturing higher levels of input information and better grasping the real data distribution, while significantly reducing the overall number of parameters.

\bibliographystyle{spmpsci}
\bibliography{QGANrefs}

\begin{thebibliography}{10}
\providecommand{\url}[1]{{#1}}
\providecommand{\urlprefix}{URL }
\expandafter\ifx\csname urlstyle\endcsname\relax
  \providecommand{\doi}[1]{DOI~\discretionary{}{}{}#1}\else
  \providecommand{\doi}{DOI~\discretionary{}{}{}\begingroup
  \urlstyle{rm}\Url}\fi

\bibitem{Arjovsky2017WassersteinG}
Arjovsky, M., Chintala, S., Bottou, L.: Wasserstein {GAN}.
\newblock arXiv preprint: arXiv:1701.07875v3  (2017)

\bibitem{Brock2019LargeSG}
Brock, A., Donahue, J., Simonyan, K.: Large scale {GAN} training for high
  fidelity natural image synthesis.
\newblock Int. Conf. on Learning Representations ({ICLR})  (2019)

\bibitem{Chen2019SelfSupervisedGV}
Chen, T., Zhai, X., Ritter, M., Lucic, M., Houlsby, N.: Self-supervised {GAN}s
  via auxiliary rotation loss.
\newblock In: {IEEE/CVF} Int. Conf. on Computer Vision and Pattern Recognition
  ({CVPR}), pp. 12146--12155 (2019)

\bibitem{TookSIGPRO2011}
Cheong~Took, C., Mandic, D.P.: Augmented second-order statistics of quaternion
  random signals.
\newblock Signal Process. \textbf{91}(2), 214--224 (2011)

\bibitem{Chernov1995}
Chernov, V.: Discrete orthogonal transforms with data representation in
  composition algebras.
\newblock Proc. Scandinavian Conf. on Image Analysis pp. 357--364 (1995)

\bibitem{ComminielloICASSP2019a}
Comminiello, D., Lella, M., Scardapane, S., Uncini, A.: Quaternion
  convolutional neural networks for detection and localization of 3{D} sound
  events.
\newblock In: {IEEE} Int. Conf. on Acoust., Speech and Signal Process.
  ({ICASSP}), pp. 8533--8537. Brighton, UK (2019)

\bibitem{ELL2007137}
Ell, T.A., Sangwine, S.J.: Quaternion involutions and anti-involutions.
\newblock Comput. Math. Appl. \textbf{53}(1), 137--143 (2007)

\bibitem{GaudetIJCNN2018}
Gaudet, C., Maida, A.: Deep quaternion networks.
\newblock In: {IEEE} Int. Joint Conf. on Neural Netw. ({IJCNN}). Rio de
  Janeiro, Brazil (2018)

\bibitem{GlorotInit2010}
Glorot, X., Bengio, Y.: Understanding the difficulty of training deep
  feedforward neural networks.
\newblock In: Int. Conf. on artificial intelligence and statistics, pp.
  249--256 (2010)

\bibitem{GoodfellowNIPS2014}
Goodfellow, I.J., Pouget-Abadie, J., Mirza, M., Xu, B., Warde-Farley, D.,
  Ozair, S., Courville, A., Bengio, Y.: Generative adversarial nets.
\newblock In: 27th Int. Conf. on Neural Information Processing Systems
  ({NIPS}), vol.~2, pp. 2672--2680. MIT Press, Cambridge, MA, USA (2014)

\bibitem{GoukLipNN2021}
Gouk, H., Frank, E., Pfahringer, B., Cree, M.J.: Regularisation of neural
  networks by enforcing {L}ipschitz continuity.
\newblock Mach. Learn. \textbf{110}(2), 393--416 (2021)

\bibitem{Grassucci2021Entropy}
Grassucci, E., Comminiello, D., Uncini, A.: An information-theoretic
  perspective on proper quaternion variational autoencoders.
\newblock Entropy \textbf{23}(7) (2021)

\bibitem{GrassucciICASSP2021}
Grassucci, E., Comminiello, D., Uncini, A.: A quaternion-valued variational
  autoencoder.
\newblock In: IEEE Int. Conf. on Acoust., Speech and Signal Process.
  ({ICASSP}). Toronto, Canada (2021)

\bibitem{GrassucciFlexGAN2021}
Grassucci, E., Scardapane, S., Comminiello, D., Uncini, A.: Flexible generative
  adversarial networks with non-parametric activation functions.
\newblock In: Progress in Artificial Intelligence and Neural Systems, vol. 184.
  Smart Innovation, Systems and Technologies, Springer (2021)

\bibitem{Gui2020ARO}
Gui, J., Sun, Z., Wen, Y., Tao, D., Ye, J.p.: A review on generative
  adversarial networks: Algorithms, theory, and applications.
\newblock arXiv preprint: arXiv:2001.06937v1  (2020)

\bibitem{GulrajaniNIPS2017}
Gulrajani, I., Ahmed, F., Arjovsky, M., Dumoulin, V., Courville, A.C.: Improved
  training of {W}asserstein {GAN}s.
\newblock In: Advances in Neural Information Processing Systems ({NIPS}) (2017)

\bibitem{HeInit2015}
He, K., Zhang, X., Ren, S., Sun, J.: Delving deep into rectifiers: Surpassing
  human-level performance on imagenet classification.
\newblock In: {IEEE/CVF} Int. Conf. on Computer Vision and Pattern Recognition
  ({CVPR}), pp. 1026--1034 (2015)

\bibitem{HeuselFID2017}
Heusel, M., Ramsauer, H., Unterthiner, T., Nessler, B., Hochreiter, S.: {GAN}s
  trained by a two time-scale update rule converge to a local {N}ash
  equilibrium.
\newblock In: Neural Information Processing Systems ({NIPS}), pp. 6626--6637
  (2017)

\bibitem{HoffmanAlgebraNets2020}
Hoffmann, J., Schmitt, S., Osindero, S., Simonyan, K., Elsen, E.:
  Algebra{N}ets.
\newblock arXiv preprint: arXiv:2006.07360v2  (2020)

\bibitem{IoffeBN2015}
Ioffe, S., Szegedy, C.: Batch normalization: Accelerating deep network training
  by reducing internal covariate shift.
\newblock In: Int. Conf. on Machine Learning ({ICML}), p. 448–456. JMLR.org
  (2015)

\bibitem{KarrasPG2018}
Karras, T., Aila, T., Laine, S., Lehtinen, J.: Progressive growing of {GAN}s
  for improved quality, stability, and variation.
\newblock In: Int. Conf. on Learning Representations ({ICLR}) (2018)

\bibitem{KarrasStyleGen2019}
Karras, T., Laine, S., Aila, T.: A style-based generator architecture for
  generative adversarial networks.
\newblock In: {IEEE} Conf. on Computer Vision and Pattern Recognition, {CVPR},
  pp. 4401--4410. Computer Vision Foundation / {IEEE} (2019)

\bibitem{KarrasSG22020}
Karras, T., Laine, S., Aittala, M., Hellsten, J., Lehtinen, J., Aila, T.:
  Analyzing and improving the image quality of stylegan.
\newblock In: 2020 {IEEE/CVF} Conf. on Computer Vision and Pattern Recognition
  ({CVPR}), pp. 8107--8116. {IEEE} (2020)

\bibitem{KingmaARXIV2014}
Kingma, D.P., Welling, M.: Auto-encoding variational {B}ayes.
\newblock arXiv Preprint: arXiv:1312.6114v10 pp. 1--14 (2014)

\bibitem{Kurach2019ALS}
Kurach, K., Lucic, M., Zhai, X., Michalski, M., Gelly, S.: A large-scale study
  on regularization and normalization in {GAN}s.
\newblock In: Int. Conf. on Machine Learning ({ICML}) (2019)

\bibitem{Miyato2018SpectralNF}
Miyato, T., Kataoka, T., Koyama, M., Yoshida, Y.: Spectral normalization for
  generative adversarial networks.
\newblock arXiv preprint: arXiv:1802.05957v1  (2018)

\bibitem{ParcolletICASSP2019a}
Parcollet, T., Morchid, M., Linar\`es, G.: Quaternion convolutional neural
  networks for heterogeneous image processing.
\newblock In: {IEEE} Int. Conf. on Acoust., Speech and Signal Process.
  ({ICASSP}), pp. 8514--8518. Brighton, UK (2019)

\bibitem{ParcolletAIR2019}
Parcollet, T., Morchid, M., Linar\`es, G.: A survey of quaternion neural
  networks.
\newblock Artif. Intell. Rev.  (2019)

\bibitem{ParcolletICLR2019}
Parcollet, T., Ravanelli, M., Morchid, M., Linar\`es, G., Trabelsi, C.,
  De~Mori, R., Bengio, Y.: Quaternion recurrent neural networks.
\newblock In: Int. Conf. on Learning Representations ({ICLR}), pp. 1--19. New
  Orleans, LA (2019)

\bibitem{Radford2016UnsupervisedRL}
Radford, A., Metz, L., Chintala, S.: Unsupervised representation learning with
  deep convolutional generative adversarial networks.
\newblock arXiv preprint: arXiv:1511.06434v2  (2016)

\bibitem{SalimansIS2016}
Salimans, T., Goodfellow, I.J., Zaremba, W., Cheung, V., Radford, A., Chen, X.:
  Improved techniques for training {GAN}s.
\newblock In: Neural Information Processing Systems ({NIPS}), pp. 2234--2242
  (2016)

\bibitem{SchmidhuberMIT1991}
Schmidhuber, J.: A possibility for implementing curiosity and boredom in
  model-building neural controllers.
\newblock In: Proc. of the First Int. Conf. on Simulation of Adaptive Behavior
  on From Animals to Animats, pp. 222–--227. MIT Press, Cambridge, MA, USA
  (1991)

\bibitem{SchmidhuberNEUNET2020}
Schmidhuber, J.: Generative adversarial networks are special cases of
  artificial curiosity (1990) and also closely related to predictability
  minimization (1991).
\newblock Neural Networks \textbf{127}, 58--66 (2020)

\bibitem{schonfeld2021unet}
Schönfeld, E., Schiele, B., Khoreva, A.: A {U}-{N}et based discriminator for
  generative adversarial networks.
\newblock In: {IEEE/CVF} Conf. on Computer Vision and Pattern Recognition
  ({CVPR}), pp. 8207--8216 (2020)

\bibitem{Qgan2021Sfikas}
Sfikas, G., Giotis, A.P., Retsinas, G., Nikou, C.: Quaternion generative
  adversarial networks for inscription detection in byzantine monuments.
\newblock In: Pattern Recognition. {ICPR} International Workshops and
  Challenges, pp. 171--184. Springer International Publishing (2021)

\bibitem{VecchiTIT2020}
Vecchi, R., Scardapane, S., Comminiello, D., Uncini, A.: Compressing
  deep-quaternion neural networks with targeted regularisation.
\newblock CAAI Trans. Intell. Technol. \textbf{5}(3), 172--176 (2020)

\bibitem{ViaTIT2010}
V\`ia, J., Ram\`irez, D., Santamar\`ia, I.: Proper and widely linear processing
  of quaternion random vectors.
\newblock IEEE Trans. Inf. Theory \textbf{56}(7), 3502--3515 (2010)

\bibitem{Ward1997}
Ward, J.P.: Quaternions and Caley Numbers. Algebra ans Applications,
  \emph{Mathematics and Its Applications}, vol. 403.
\newblock Kluwer Academic Publishers (1997)

\bibitem{QilinQCNN2019}
Yin, Q., Wang, J., Luo, X., Zhai, J., Jha, S.K., Shi, Y.: Quaternion
  convolutional neural network for color image classification and forensics.
\newblock {IEEE} Access \textbf{7}, 20293--20301 (2019)

\bibitem{ZhangSAGAN2019}
Zhang, H., Goodfellow, I.J., Metaxas, D.N., Odena, A.: Self-attention
  generative adversarial networks.
\newblock In: Int. Conf. on Machine Learning ({ICML}), \emph{Proceedings of
  Machine Learning Research}, vol.~97, pp. 7354--7363. {PMLR} (2019)

\bibitem{ZhangConsRegGAN2020}
Zhang, H., Zhang, Z., Odena, A., Lee, H.: Consistency regularization for
  generative adversarial networks.
\newblock In: Int. Conf. on Machine Learning ({ICML}) (2020)

\bibitem{ZhouLipGan2019}
Zhou, Z., Liang, J., Song, Y., Yu, L., Wang, H., Zhang, W., Yu, Y., Zhang, Z.:
  Lipschitz generative adversarial nets.
\newblock In: Int. Conf. on Machine Learning ({ICML}), \emph{Proceedings of
  Machine Learning Research}, vol.~97, pp. 7584--7593. {PMLR} (2019)

\end{thebibliography}

\end{document}